\PassOptionsToPackage{compress}{natbib}
\documentclass{article}

\newcommand{\eat}[1]{}

\usepackage{latexsym}
\usepackage{amsfonts}
\usepackage{amsmath}
\usepackage{xcolor}
\usepackage{colortbl}
\usepackage{epsfig}
\usepackage{xspace}
\usepackage{graphicx}
\usepackage{subfigure}
\usepackage{paralist}
\usepackage{enumerate}

\usepackage{amssymb}
\usepackage{enumitem}
\setlist{noitemsep,parsep=0pt,partopsep=0pt, leftmargin=10pt} 
\usepackage[color,matrix,arrow,all]{xy}
\usepackage{comment}
\usepackage{booktabs}
\usepackage{balance}
\usepackage{stmaryrd}
\usepackage{pifont}
\usepackage{hhline}
\usepackage{listings}
\usepackage{array}
\usepackage{float}
\usepackage[flushleft]{threeparttable}

\usepackage{mathrsfs}
\usepackage{makecell}
\usepackage{xparse}
\usepackage{wrapfig}

\usepackage{epsfig}
\usepackage{multirow}
\usepackage{url}

\usepackage{multirow}
\usepackage{natbib}
\usepackage{graphicx}
\usepackage{subfigure}
\usepackage{svg}

\usepackage{listings}
\usepackage{framed}
\usepackage{xcolor}
\usepackage{color}
\usepackage{changepage}
\colorlet{shadecolor}{gray!20}
\usepackage{makecell}

\usepackage{algorithm}
\usepackage{algorithmic}

\definecolor{shadecolor}{RGB}{220,220,220}

\definecolor{lightgreen}{HTML}{D5E8D4}
\definecolor{lightyellow}{HTML}{fbe7cf}
\definecolor{inputcolor}{RGB}{255,139,35}
\definecolor{outputcolor}{RGB}{120,212,252}
\definecolor{embedcolor}{RGB}{254,127,156}
\definecolor{maskcolor}{RGB}{122,128,255}
\definecolor{ecolor}{RGB}{58,149,54}

\definecolor{highcolor}{RGB}{255,153,153}
\definecolor{midcolor}{RGB}{255,204,204}
\definecolor{lowcolor}{RGB}{204,229,255}

\newcommand{\cmarkbar}{\cmark\hspace{-0.6em}\rule[0.8ex]{0.6em}{0.3mm}}

\usepackage{tikz}
\usetikzlibrary{shapes,decorations.pathmorphing}
\usetikzlibrary{calc}

\usepackage{algorithm} 
\usepackage{algorithmic}  
\usepackage[algo2e]{algorithm2e} 

\usepackage[export]{adjustbox}

\definecolor{red}{RGB}{184,84,80}

\definecolor{green}{RGB}{0,128,0}

\definecolor{yellow}{RGB}{255,200,18}

\definecolor{blue}{RGB}{33,89,202}

\newcommand{\stab}{\vspace{1.2ex}\noindent}

\newcommand{\bi}{\begin{itemize}}
\newcommand{\ei}{\end{itemize}}

\newcommand{\be}{\begin{enumerate}}
\newcommand{\ee}{\end{enumerate}}
\newcommand{\beqn}{\begin{eqnarray*}}
\newcommand{\eeqn}{\end{eqnarray*}}

\newcommand{\stitle}[1]{\stab\noindent{\bf #1}}
\newcommand{\etitle}[1]{\vspace{1mm}\noindent{\underline{\em #1}}}

\newcommand{\ie}{\textit{i.e.,} \xspace}
\newcommand{\eg}{\textit{e.g.,} \xspace}

\newcommand{\nvbench}{{nvBench}\xspace}
\newcommand{\nlv}{{NLV Corpus}\xspace}

\newcommand{\model}{{Step-Text2Vis}\xspace}
\newcommand{\data}{{nvBench~2.0}\xspace}
\newcommand{\dataset}{{nvBench~2.0}\xspace}

\newcommand{\vql}{\textsc{vql}\xspace}

\newcommand{\nlq}{\textsc{text}\xspace}
\newcommand{\vis}{\textsc{vis}\xspace}
\newcommand{\sql}{\textsc{sql}\xspace}
\newcommand{\nlvis}{{\em Text2VIS}\xspace}
\newcommand{\nlsql}{{\em Text2SQL}\xspace}

\usepackage{pifont} 
\newcommand{\cmark}{\ding{51}} 
\newcommand{\xmark}{\ding{55}} 

\makeatletter
    \newcommand\figcaption{\def\@captype{figure}\caption}
    \newcommand\tabcaption{\def\@captype{table}\caption}
\makeatother

\tikzstyle{mybox} = [draw=black, fill=black!5, thick,
   rectangle, rounded corners, inner sep=0pt, inner ysep=6pt]
\tikzstyle{fancytitle} =[fill=black, text=white]

\NewDocumentCommand{\nan}{ mO{} }{\textcolor{blue}{\textsuperscript{\textit{Nan}}\textsf{\textbf{\small[#1]}}}}

\NewDocumentCommand{\yuyu}{ mO{} }{\textcolor{green}{\textsuperscript{\textit{Yuyu}}\textsf{\textbf{\small[#1]}}}}

\NewDocumentCommand{\leixian}{ mO{} }{\textcolor{blue}{\textsuperscript{\textit{Leixian}}\textsf{\textbf{\small[#1]}}}}

\NewDocumentCommand{\fanj}{ mO{} }{\textcolor{red}{\textsuperscript{\textit{Ju}}\textsf{\textbf{\small[#1]}}}}

\NewDocumentCommand{\tianqi}{ mO{} }{\textcolor{violet}{\textsuperscript{\textit{Tianqi}}\textsf{\textbf{\small[#1]}}}}

\NewDocumentCommand{\wei}{ mO{} }{\textcolor{blue}{\textsuperscript{\textit{Wei}}\textsf{\textbf{\small[#1]}}}}

\usepackage[most]{tcolorbox}

\tcbuselibrary{breakable}

\newtcolorbox{examplebox}[1]{
 enhanced,
 breakable, 
 colback=gray!10,
 colframe=gray!70,
 width=\columnwidth,
 boxrule=0.4pt,
 left=0mm,
 right=0mm,
 top=0mm,
 bottom=0mm,
 before skip balanced=0pt,
 after skip balanced=0pt,
 arc=5pt,
 fonttitle=\bfseries,
 title=#1,
 coltitle=white,  
 oversize
}

\usepackage{listings}
\usepackage{xcolor}  

\lstdefinelanguage{JSON}{
    string=[s]{"}{"},
    stringstyle=\color{red},
    comment=[l]{:},
    commentstyle=\color{blue},
    morecomment=[l]{,},
    morecomment=[l]{[},
    morecomment=[l]{]},
    morecomment=[l]{\{},
    morecomment=[l]{\}},
    keywordstyle=\color{blue},
    keywords={true, false, null}
}

\lstdefinestyle{jsonstyle}{
  language=JSON,
  basicstyle=\rmfamily\small,  
  keywordstyle=\color{blue},   
  stringstyle=\color{red},     
  showstringspaces=false,      
  commentstyle=\color{green},  
  breaklines=false,             
  numbers=none                 
}

\usepackage{soul}

\definecolor{verylightgray}{rgb}{0.95,0.95,0.95} 
\newcommand{\asprule}[1]{%
  \begingroup
  \sethlcolor{verylightgray}%
  \hl{\texttt{#1}}%
  \endgroup
}

\usepackage[dandb, final]{neurips_2025}

\usepackage[utf8]{inputenc} 
\usepackage[T1]{fontenc}    
\usepackage{hyperref}       
\usepackage{url}            
\usepackage{booktabs}       
\usepackage{amsfonts}       
\usepackage{graphicx}       
\usepackage{array} 
\newcolumntype{C}[1]{>{\centering\arraybackslash}p{#1}}
\usepackage{nicefrac}       
\usepackage{microtype}      
\usepackage{xcolor}         
\usepackage{newtxtext}
\usepackage{tabularx}
\usepackage{ragged2e} 
\usepackage{geometry} 

\title{nvBench 2.0: Resolving Ambiguity in Text-to-Visualization through Stepwise Reasoning}

\author{
  \textbf{Tianqi Luo}$^{1}$ \quad \textbf{Chuhan Huang}$^{1}$ \quad \textbf{Leixian Shen}$^{2}$ \quad \textbf{Boyan Li}$^{1}$ \quad \textbf{Shuyu Shen}$^{1}$  \quad\\
  \textbf{Wei Zeng}$^{1,2}$ \quad \textbf{Nan Tang}$^{1,2}$ \quad \textbf{Yuyu Luo}$^{1,2}$\thanks{Yuyu Luo is the corresponding author (yuyuluo@hkust-gz.edu.cn)} \\
  $^{1}$The Hong Kong University of Science and Technology (Guangzhou), Guangzhou, China \\
  $^{2}$The Hong Kong University of Science and Technology, Hong Kong SAR, China \\
}

\begin{document}

\maketitle

\begin{abstract}
Text-to-Visualization (\nlvis) enables users to create visualizations from natural language queries, making data insights more accessible. However, \nlvis faces challenges in interpreting ambiguous queries, as users often express their visualization needs in imprecise language. To address this challenge, we introduce \dataset, a new benchmark designed to evaluate \nlvis systems in scenarios involving ambiguous queries.
\dataset includes 7,878 natural language queries and 24,076 corresponding visualizations, derived from 780 tables across 153 domains.
It is built using a controlled ambiguity-injection pipeline that generates ambiguous queries through a reverse-generation workflow. By starting with unambiguous seed visualizations and selectively injecting ambiguities, the pipeline yields multiple valid interpretations for each query, with each ambiguous query traceable to its corresponding visualization through step-wise reasoning paths.
We evaluate various Large Language Models (LLMs) on their ability to perform ambiguous \nlvis tasks using \dataset. We also propose \model, an LLM-based model trained on \dataset, which enhances performance in ambiguous scenarios through step-wise preference optimization. Our results show that \model outperforms all baselines, setting a new state-of-the-art for ambiguous \nlvis tasks.
Our source code and data are available at \url{https://nvbench2.github.io/}.
\end{abstract}


\section{Introduction}
\label{sec:intro}

Text-to-Visualization (\nlvis) democratizes data exploration and analysis by enabling users to generate Visualizations (\vis) from text queries~\cite{NLISurvey, Voigt2022, deepeye_keyword,deepeye_tkde,deepeye}.
While recent advances in Large Language Models (LLMs)~\cite{Chen2024a,Author2023,ChartInsights,nl2sql_survey,aflow,DBLP:journals/corr/abs-2506-09507,DBLP:journals/corr/abs-2508-02124, zhu2025surveydataagent, li2025deepeyesql} have significantly enhanced translation accuracy, they struggle with a fundamental challenge: \textit{text query ambiguity}—a single query often maps to multiple valid visualizations, each representing a different interpretation of the user's intent~\cite{DataTone2015,Hoque2018,vistalk,ambiQT2023,nl2sql360, NL4DV2021, sah2024nl4dvllm, zhang2021tadoc}.

\begin{figure}[t!]
\centering
\includegraphics[width=\columnwidth]{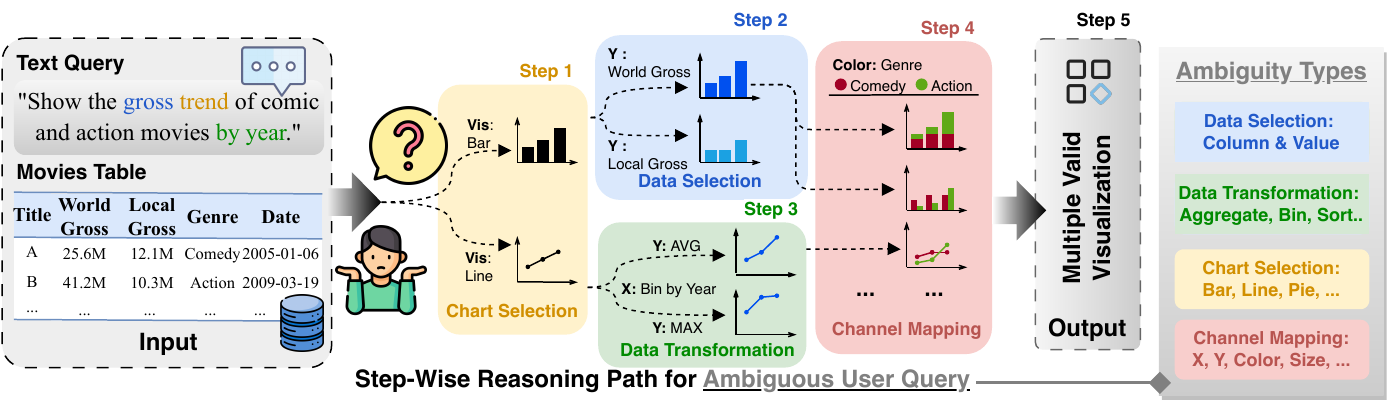}
\vspace{-1.5em}
\caption{Example of reasoning appropriate visualizations from an ambiguous query.}
\label{fig:motivation_example_1}
\vspace{-1.em}
\end{figure}

In \nlvis, there are two layers of ambiguity: the \textbf{data layer}, which governs how a query selects, filters, and transforms data, and the \textbf{visualization layer}, which determines how the data is visually represented. 
For example, in Figure~\ref{fig:motivation_example_1}, the text query ``{\tt Show the \textcolor{blue}{gross} \textcolor{yellow}{trend} of comedy and action movies \textcolor{green}{by year}}'' appears straightforward but contains multiple ambiguities.
At the data layer, ``{\tt \textcolor{blue}{gross}}'' could mean either {\tt World\_Gross} or {\tt Local\_Gross} column,
while ``{\tt comedy and action}'' implicitly requires filtering the column \texttt{Genre}.
Moreover, ``{\tt \textcolor{green}{by year}}'' implies temporal binning on {\tt Date} with aggregation on ``{\tt \textcolor{blue}{gross}}'', neither of which is explicitly specified. 
At the visualization layer, ``{\tt \textcolor{yellow}{trend}}'' might suggest a bar or line chart, with the {\tt \textcolor{red}{x-channel}} representing the binned temporal data, the {\tt \textcolor{red}{y-channel}} showing the aggregated ``{\tt \textcolor{blue}{gross}}'' values, and the implicit column {\tt Genre} mapped to the {\tt \textcolor{red}{color-channel}}, where comedy and action are represented by distinct colors in the visualization. This example highlights how ambiguities at both the data and visualization layers interact, complicating the mapping from text queries to visualizations. 

\textbf{Existing Benchmarks and Their Limitations.}
Although several benchmarks for the \nlvis task exist~\cite{VL2NL2024,nvBench2021,Chen2024a,nlv2021,QUDA2020,dail-nvbench2023,nvBench_dataset}, as shown in Table~\ref{tab:compare}, none explicitly evaluates how the \nlvis systems handle ambiguity.
In fact, existing efforts~\cite{nvBench2021,dail-nvbench2023,Chen2024a} often overlook this issue by adhering to the \textit{single-correct-answer} paradigm, where each text query maps to exactly one valid visualization.
For example, nvBench~\cite{nvBench2021} maps a text query to a unique visualization, ignoring more than 60\% of real-world ambiguous cases~\cite{nlv2021}.
Similarly, Dial-NVBench~\cite{dail-nvbench2023} supports multi-turn clarification but assumes that the final query is well-specified, which sidesteps the inherent ambiguities in user intents.

This narrow focus leaves a critical gap in the push to advance \nlvis systems. \textit{How can we evaluate and improve their ability to generate valid visualizations from ambiguous text queries?}

\textbf{Design Considerations.}
To address this challenge, a benchmark is needed that tests \nlvis solutions on handling ambiguous text queries, recognizing multiple valid interpretations, and providing an appropriate set of answers. This benchmark should include diverse ambiguous queries, multiple valid outputs, reasoning paths explaining the ambiguity, and broad domain coverage.

\textbf{Our Proposal.}
To fill this gap, we propose \dataset, the first benchmark for generating visualizations from ambiguous text queries (\ie {the ambiguous \nlvis task}). This dataset provides a robust foundation for evaluating \nlvis solutions in scenarios where text query ambiguity is a key challenge.
\dataset includes 7,878 text queries and 24,076 corresponding visualizations, derived from 780 tables across 153 domains. It meets the design considerations through a controllable ambiguity-injected \nlvis data synthesis pipeline. This pipeline uses a reverse-generation workflow, starting with a seed visualization and injecting ambiguity to create multiple valid interpretations. We then generate an ambiguous text query for each set of modified visualizations, incorporating injected ambiguity. 
The pipeline’s transparency allows tracking of how interpretations lead to distinct visualizations, supported by reasoning paths detailing the ambiguity resolution process. 
This traceability enables researchers to assess the effectiveness and interpretability of ambiguity resolution, ensuring an accurate and explainable process.

\textbf{Contributions.} Our main contributions are summarized as follows:

\begin{itemize}
    \item \textbf{Ambiguity-Injected Data Synthesizer.} We develop a \nlvis data synthesizer that generates ambiguous data by selectively injecting ambiguities into seed visualizations, yielding multiple valid answers for each text query while providing step-wise disambiguation reasoning paths. (Section~\ref{sec:synthesizer})
    
    \item \textbf{\dataset Benchmark.} We present \dataset, the first benchmark designed for the ambiguous \nlvis task. It contains 7,878 text queries and 24,076 corresponding visualizations, derived from 780 tables across 153 domains. Each \nlvis sample is paired with a disambiguation reasoning path, providing clear explanations of how the ambiguity is resolved and ensuring the interpretability of the ambiguity resolution process. (Section~\ref{sec:characteristics})
    
     \item \textbf{\model for Ambiguous \nlvis Tasks.}
      We propose \model, an LLM-based model trained on \dataset. By leveraging step-wise preference optimization and the provided reasoning paths, \model model achieves the highest F1@3 (81.50\%) and F1@5 (80.88\%), outperforming prompting GPT-4o by 22.54\% and 21.85\%, respectively. (Section~\ref{sec:model})
     
    \item \textbf{Extensive Evaluation.} 
    We conduct comprehensive experiments to validate the effectiveness of \dataset for training and evaluating \nlvis systems under ambiguity. Our findings reveal the limitations of existing models when faced with ambiguous queries while demonstrating that the \model model outperforms baseline approaches and achieves state-of-the-art performance in ambiguous \nlvis tasks.
    (Section~\ref{sec:exp})
\end{itemize}


\begin{table*}[t!]
\centering
\caption{Comparison of \nlvis benchmarks.}
\label{tab:compare}
\small
\renewcommand\arraystretch{0.7}
\setlength{\tabcolsep}{1.5mm}{
\begin{tabular}{cccccccc}
\toprule
\multirow{2}{*}{Datasets} &
  \multirow{2}{*}{\#-Tables} &
  \multicolumn{2}{c}{\#-Samples} &
  \multirow{2}{*}{\begin{tabular}[c]{@{}c@{}} \nlq $\rightarrow$ \vis\\ Mapping\end{tabular}} &
  \multirow{2}{*}{\begin{tabular}[c]{@{}c@{}}\nlq \\Ambiguity \end{tabular}} &
  \multirow{2}{*}{\begin{tabular}[c]{@{}c@{}}Reasoning\\ Paths\end{tabular}} &
  \multirow{2}{*}{\begin{tabular}[c]{@{}c@{}}\nlq \\ Generation\end{tabular}} \\ \cline{3-4} 
 &
   &
  \multicolumn{1}{c}{\#-\vis} &
  \multicolumn{1}{c}{\begin{tabular}[c]{@{}c@{}}\#-\nlq\end{tabular}} &
   &
   \\ 
\specialrule{.2pt}{2pt}{2pt} 
Quda~\cite{QUDA2020}  & 36   & \multicolumn{1}{c}{-}    & \multicolumn{1}{c}{14035}   &  \multicolumn{1}{c}{-} &  \cmarkbar & \xmark & Human-based      \\ 
NLV~\cite{nlv2021}  & 3   & \multicolumn{1}{c}{30}    & \multicolumn{1}{c}{814}   &  \multicolumn{1}{c}{$n \rightarrow 1$} & \cmarkbar & \xmark & Human-based      \\ 
Dial-nvBench~\cite{dail-nvbench2023}     & 780  & \multicolumn{1}{c}{7247}  & \multicolumn{1}{c}{124449} &  \multicolumn{1}{c}{$n \rightarrow 1$} & \xmark & \xmark & Rule-based \\  
VL2NL~\cite{VL2NL2024}       &  1981   & \multicolumn{1}{c}{1981}  & \multicolumn{1}{c}{3962}  &  \multicolumn{1}{c}{$1 \rightarrow 1$} & \cmarkbar & \xmark & LLM-based \\ 
VisEval~\cite{Chen2024a}     &  748   & \multicolumn{1}{c}{2524}  & \multicolumn{1}{c}{1150}  &  \multicolumn{1}{c}{$1 \rightarrow 1$} & \xmark & \xmark & LLM-based \\ 
nvBench~\cite{nvBench2021}     & 780 & \multicolumn{1}{c}{7247}  & \multicolumn{1}{c}{25750} &  \multicolumn{1}{c}{$1 \rightarrow 1$} & \xmark & \xmark & Rule-based \\ 
\specialrule{.2pt}{2pt}{2pt} 
\textbf{\dataset} & 780 & \multicolumn{1}{c}{{24076}} & \multicolumn{1}{c}{7878}  & \multicolumn{1}{c}{$1 \rightarrow n$} & {\cmark} & {\cmark} & LLM-based \\
\bottomrule
\end{tabular}}
\end{table*}

\section{\dataset}
\label{sec:pre}


In this section, we will first elaborate on how to develop \dataset with an Ambiguity-Injected Data Synthesizer (Section~\ref{sec:synthesizer}), and then describe the characteristics of \dataset (Section~\ref{sec:characteristics}).


\subsection{Ambiguity-Injected \nlvis Data Synthesizer}
\label{sec:synthesizer}

\begin{figure*}[t!]
	\centering
\includegraphics[width=\textwidth]{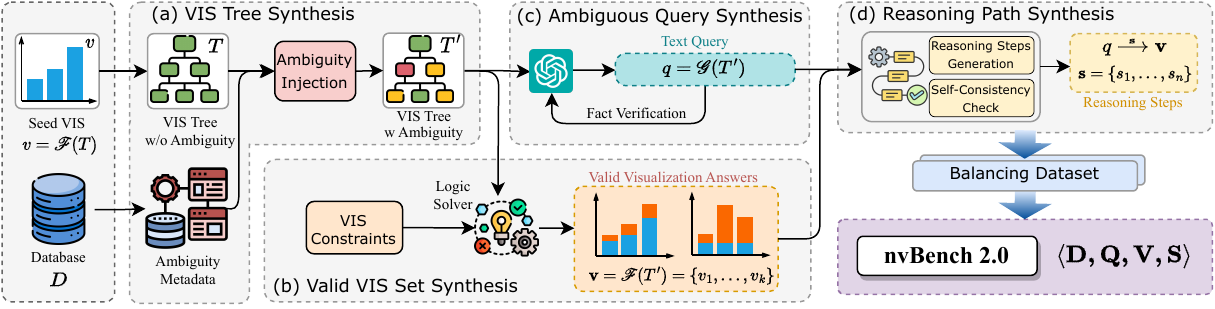}
\vspace{-1em}
	\caption{The Pipeline for Synthesizing \data.}
	\label{fig:pipeline}
    \vspace{-.5em}
\end{figure*}

Figure~\ref{fig:pipeline} provides a high-level overview of our \emph{Ambiguity-Injected \nlvis Data Synthesizer}. The pipeline begins with a data table $D$ and an unambiguous seed \vis $v$, from which a \vis tree $T$ (without ambiguity) is derived. Through a systematic ambiguity injection process, this tree is transformed into $T'$ (with ambiguity), serving as intermediate products in the workflow. Subsequently, the pipeline generates an ambiguous text query $q$ alongside a corresponding set of valid \vis $\mathbf{v}=\{v_1, \dots, v_k\}$, while also producing step-wise reasoning paths $\mathbf{s}=\{s_1, \dots, s_k\}$ for each valid \vis. The final output of the pipeline is \dataset, structured in the form $\langle D, Q, V, S \rangle$.
Next, we will go through our pipeline step by step.


\textbf{Data Preparation.}
To construct \dataset, we obtained data tables from {nvBench~1.0}~\cite{nvBench2021} and BIRD~\cite{BIRD2023}, ensuring broad coverage of the domain and relevance in the real world. We retained select seed visualizations from \cite{nvBench2021} while introducing new ones to address the limited variety of chart types. Novel chart types, such as boxplots and heatmaps, and additional encoding channels, like "size" (enabling scatterplots to transition to bubble charts), were incorporated to enhance visual encoding diversity and expressiveness.


\textbf{Step 1: Ambiguity-aware VIS Tree Synthesis.}
We start the process by building an initial \vis tree $ T $ from a seed \vis $ v $. The tree $ T $, structured as an Abstract Syntax Tree (AST) based on \vis grammar, encodes components such as data mappings, mark types, and encoding channels. Each node in $ T $ represents a component in \vis and whether it has ambiguity.
We then extract ambiguity metadata from the database $D$ using a structured knowledge graph (KG). This involves a KG-driven semantic alias identification process and an LLM-driven refinement process. Metadata systematically categorizes semantic ambiguities within the table schema to guide the creation of ambiguity-aware \vis trees.

Subsequently, we transform $ T $ into an ambiguity-aware tree $ T' $ through a controlled ambiguity injection process. This involves two operations: (1) \emph{injecting ambiguous nodes} to add semantically ambiguous components, (2) \emph{injecting implicit nodes} to replace fixed components with a blank placeholder, or to add unspecified but essential components for the altered \vis tree. 
This injection process ensures that $ T' $ can branch into multiple valid interpretations, capturing the full spectrum of possible outcomes for an ambiguous text query.

\textbf{Step 2: Valid VIS Set Synthesis.}
The partially ambiguous \vis tree $T'$  is processed through an Answer Set Programming (ASP) solver~\cite{gebser2019multi}, which applies grammar constraints to transform the ambiguous tree into a resolved \vis set $\mathbf{v} = \{v_1, \ldots, v_k\}$. The number of resulting \vis, $k = |\mathbf{v}|$, indicates the ambiguity level—how many distinct interpretations the solver deems valid for given $T$.

The completeness of a valid answer set is ensured by the ASP solver's exhaustive enumeration of all stable models satisfying the encoded \vis constraints through its declarative logic programming framework~\cite{gebser2019multi}.
This guarantees that all possible answers consistent with the input text query and grammar constraints are generated, providing a comprehensive set of solutions for ambiguous queries. 
Please refer to Appendix~\ref{sub:step2} for more details.

\textbf{Step 3: Ambiguous Query Synthesis.}
We leverage an LLM-based {\tt Query Generator} to synthesize an ambiguous text query $q$ for each modified \vis tree $T'$, incorporating the newly introduced ambiguities into a single query. This approach ensures that every synthesized valid answer in set $\mathbf{v}=\{v_1, \dots, v_k\}$ faithfully represents the ambiguous intents of the query. Finally, an LLM-based {\tt Query Verifier} checks consistency, confirming that the final text query accurately reflects all valid answers.
Please refer to Section~\ref{sub:step3} in the Appendix for more details.

\textbf{Step 4: Ambiguity-resolved Reasoning Paths Synthesis.}
Finally, we generate \emph{stepwise disambiguation reasoning paths} to guide the resolution of each ambiguity in producing every valid \vis. These paths are built using an LLM with an automated self-consistency validation mechanism to ensure accuracy.
By systematically extracting and articulating the discrete reasoning steps from the initial text query $q$ to the set of \vis $\mathbf{v}$, we provide a transparent and comprehensive explanation of how each query maps to its corresponding \vis outcomes. Details are provided in Section~\ref{sub:step4}.

\textbf{Dataset Balancing and Quality Control.}
We implement several strategies to ensure the high quality of our \dataset.
(1) \emph{Ambiguity Level Regulation}: We constrain the ambiguity level to $k \leq 5$, ensuring that each retained sample illustrates a meaningfully different way of interpreting the partially ambiguous tree.  
(2) \emph{Visualization Diversity}: From a large pool of randomly generated seed visualizations and ambiguity-injected trees, we compute pairwise distances and select the most diverse subset to enhance the variety of visualizations.  
(3) \emph{Query Verification}: In the query synthesis process, although GPT-4o effectively captures most ambiguous nodes in $T'$, it introduces unintended facts in approximately 5\% of generated queries. We implement a fact verification process to identify and refine these unwanted intents. We then conduct manual reviews by two postgraduate students to re-check the identified queries.
This pipeline ensures that each ambiguous aspect in \vis tree synthesis phase is clearly reflected in the final mapping of \nlvis, allowing researchers and practitioners to evaluate how effectively models handle and explain different interpretations.

\subsection{\dataset Characteristics}
\label{sec:characteristics}


This section describes the key characteristics of \dataset, focusing on the interaction of ambiguity types, levels, patterns, query styles, and visualizations.

\begin{figure}[t!]
    \centering
    \includegraphics[width=\linewidth]{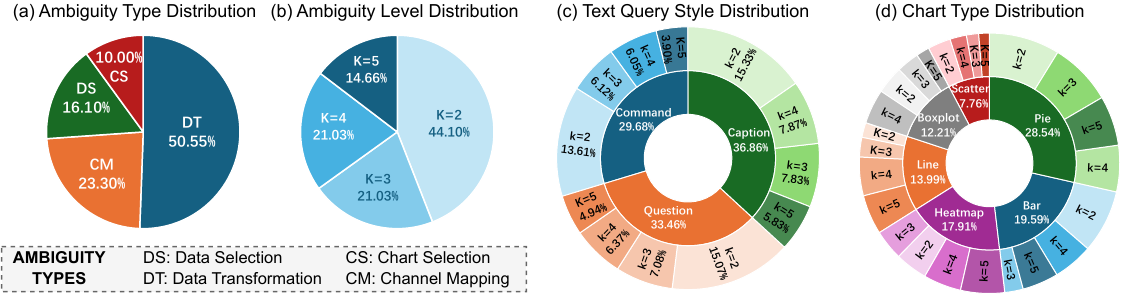}
    \vspace{-1em}
    \caption{Key Statistics of \data.}
    \label{fig:data_statistic}
\end{figure}

\textbf{Ambiguity Types and Levels.}
An important contribution of \dataset is the systematic introduction of controlled ambiguity types and ambiguity levels. 
Figure~\ref{fig:data_statistic} (a) categorizes ambiguity by type: Data Transformation (DT) ambiguities are most prevalent (50.55\%), followed by Channel Mapping (CM) ambiguities (23.30\%), with Data Selection (DS) and Chart Type Selection (CT) ambiguities represented by 16.10\% and 10.00\%, respectively. 
As shown in Figure~\ref{fig:data_statistic} (b), the majority of samples (44.10\%) have an ambiguity level of 2, indicating that two valid visualizations exist for each text query. \dataset also contains a substantial number of samples with ambiguity levels of 3, 4, and 5, enabling a thorough evaluation of systems under increasingly complex ambiguous scenarios.

\textbf{Ambiguity Combination Patterns.}
Since multiple ambiguity types can occur in a single data sample, Table~\ref{tab:ambigiuty_pattern} shows the most frequent ambiguity combination patterns.
The most common pattern is CM+DT (2,190 instances), followed by CM+DS (486), while more complex multi-category ambiguities are less frequent. Channel Mapping and Data transformation ambiguities often co-occur with other ambiguity types. This distribution underscores the challenge of resolving overlapping ambiguity types in \vis generation tasks.

\textbf{Text Queries.}
Figure~\ref{fig:data_statistic}(c) presents the query style distribution. We balance the distribution to follow the observation in work~\cite{nlv2021}, where command-based queries are most frequent, question-based and caption-like queries are also useful but less frequent. 
Appendix~\ref{apdx:dataset} provides a detailed breakdown of query styles across different chart types, along with word count statistics. 

\textbf{Visualizations.}
Figure~\ref{fig:data_statistic}(d) illustrates the distribution of the chart types. \dataset comprises six chart types, with Bar and Pie being the most prevalent, aligning their common use in the real world. 
Other chart types cater to more specialized analytical purposes, aligning with typical \vis practices. Notably, the ambiguity levels are distributed similarly across all chart types, ensuring a well-balanced data distribution to evaluate \nlvis systems under varying degrees of ambiguity.

% ...

\begin{table}[t!]
\centering
\caption{Statistics of Ambiguity Combination Patterns}
\small
\vspace{-0.7em}
\begin{tabular}{|C{3cm}|C{2cm}||C{3cm}|C{2cm}|}
\hline
\textbf{Pattern} & \textbf{Count} & 
\textbf{Pattern} & \textbf{Count} \\
\hline\hline
CM+DT & 2190 & CM+DS & 486  \\
\hline
CM+DS+DT & 364  & CM+CS+DT    & 171  \\
\hline
CM+CS & 158  & CS+DT    & 34 \\
\hline
\end{tabular}
\label{tab:ambigiuty_pattern}
\end{table}


\section{\model for Ambiguous \nlvis}
\label{sec:model}


In this section, we present \model, a new model for the ambiguous \nlvis task. \model addresses ambiguity by incorporating a step-wise reasoning process, as detailed in Section~\ref{sub:step4}, and leveraging the rich step-wise data provided by \data. Built on base LLMs, \model is fine-tuned on \data using a pipeline that aligns its outputs with the dataset's reasoning paths via supervised fine-tuning and step-wise preference optimization (Step-DPO)~\cite{StepDPO2024}.



\subsection{Preference Optimization with Step-DPO}
\label{sub:step-dpo}

Previous \nlvis methods have typically employed either prompting LLMs~\cite{Wang2023c} or fine-tuning LLMs~\cite{dail-nvbench2023}, where the LLM is directly tasked with generating the final \vis definition based on \nlq and table schema information. 

Recently, process supervision paradigms~\cite{DBLP:conf/iclr/LightmanKBEBLLS24} and preference optimization techniques~\cite{StepDPO2024} have demonstrated significant advancements across various domain tasks. A pivotal aspect in validating the effectiveness of \data is determining how to leverage the step-wise disambiguation reasoning paths within the \data dataset to provide process supervision and enhance model performance. 
Consequently, we adopt the Step-DPO~\cite{StepDPO2024}, which utilizes step-wise paired correct and incorrect samples for preference optimization, thereby delivering rich process supervision signals to the model and fostering improved accuracy at each step. 

Formally, we define an input prompt $x$ and an \vis answer $y$, where $x$ includes \nlq and table schema, and $y$ can be represented as $s_{1} \oplus \cdots \oplus s_{n}$, where $s_{i}$ denotes the $i$-th reasoning step defined in Section~\ref{sub:step4}. Given the input $x$ and a sequence of correct preceding reasoning steps $s_{1 \sim k-1}=s_{1} \oplus \cdots \oplus s_{k-1}$, Step-DPO aims to maximize the probability of the correct next reasoning step $s_{win}$ and minimize the probability of the incorrect one $s_{lose}$. 
This objective can be formulated as:


\begin{equation}
\small
    \begin{aligned}
        \mathcal{L}(\theta) &= -\mathbb{E}_{(x, s_{1 \sim k-1}, s_{win}, s_{lose}) \sim D_p} \\ &\bigg[
        \log \sigma \bigg( \beta \log \frac{\pi_{\theta}(s_{win} | x, s_{1 \sim k-1})}{\pi_{ref}(s_{win} | x, s_{1 \sim k-1})} 
        - \beta \log \frac{\pi_{\theta}(s_{lose} | x, s_{1 \sim k-1})}{\pi_{ref}(s_{lose} | x, s_{1 \sim k-1})} \bigg) \bigg]
    \end{aligned}
\end{equation}
where $D_p$ represents a step-wise preference dataset. $\pi_{\theta}(\cdot|x, s_{1 \sim k-1})$ denotes the policy model to be optimized, while $\pi_{ref}(\cdot|x, s_{1 \sim k-1})$ refers to the reference model, which remains unchanged during the training process. The hyperparameter $\beta$ controls the divergence between the optimized policy and the reference model.


\subsection{Cold-start with Supervised Fine-tuning}

Prior studies, such as those employing Chain-of-Thought (CoT)~\cite{cot} prompting, have demonstrated the capability of LLMs to engage in step-wise reasoning through the utilization of simple ``{\tt think step-by-step}'' instructions. However, under this paradigm, the planning of steps and the format of output are indiscriminate. This poses challenges in the precise extraction of answers corresponding to each individual step, and consequently, impedes the accurate alignment with the step-wise data provided within the \data dataset for the purpose of validating step-level correctness. To address this limitation, we use \data training set and employ Supervised Fine-Tuning (SFT) as a cold-start mechanism to facilitate the LLM's learning of our predefined step-wise output format. The specific training setup and prompt templates are in Appendix~\ref{apdx:exp}.

\subsection{Step-wise Preference Data Construction}

A crucial aspect of Step-DPO is the acquisition of a step-wise preference dataset. As described in Section~\ref{sub:step-dpo}, our \data dataset contains step-wise ground-truth. Therefore, we adopt an online data collection strategy. Initially, we utilize a model that has undergone Supervised Fine-Tuning (SFT) cold-start to perform inference on the \data development set, yielding $D_{0} = \{(x, \hat{y})\}$, where $\hat{y}$ represents the model's step-wise output, expressible as $\hat{s}_{1} \oplus \cdots \oplus \hat{s}_{n}$. Subsequently, we conduct a step-wise evaluation comparing $\hat{y}$ with the ground-truth $y$, verifying the correctness of each step until the identification of the first error, and recording its corresponding step number $k$. We designate the erroneous step $\hat{s}_{k}$ as the incorrect reasoning step $s_{lose}$, and the ground-truth step $s_{k}$ as the correct reasoning step $s_{win}$. The construction of the preference dataset $D_{p} = \{(x, \hat{s}_{1 \sim k-1}, s_{win}, s_{lose}\}$ is then readily achieved through the integration of input $x$ and previous reasoning steps $\hat{s}_{1 \sim k-1}$.


\section{Experiments}
\label{sec:exp}



In our experiments, we aim to answer two fundamental questions about ambiguous \nlvis tasks. 
First, how effectively do different approaches—including state-of-the-art LLMs and our proposed \model—handle \vis generation from text queries with varying levels of ambiguity? 
Second, what impact does step-wise reasoning have on performance across different chart types and ambiguity scenarios compared to direct generation approaches? To address these questions, we designed a comprehensive evaluation framework comparing different methods with or without stepwise reasoning. We assess performance using standard information retrieval metrics across multiple levels of ambiguity.

\subsection{Experimental Setup}

\textbf{Datasets.}
We use \data for our experiments and randomly divide the data set into training, development, and testing sets in a ratio of 80\%, 10\%, and 10\%, containing 6377, 750, and 751 samples, respectively. 

\textbf{Methods.}
We evaluate the performance on ambiguous \nlvis tasks using both prompting-based and fine-tuning-based methods with \data. The primary goal is to assess the model's ability to generate diverse and semantically accurate visualizations in response to ambiguous text queries. Please refer to Section~\ref{apdx:exp} in the Appendix for more details.

\etitle{Prompting-based Methods.}
We evaluate two prompting strategies: Direct Prompting and Step Prompting. In Direct Prompting, the model receives structured \textit{Data Schema Information} and a \textit{text query} as input, subsequently generating 1-5 distinct visualizations to cover possible interpretations of the ambiguous query. This strategy is applied to the \textbf{GPT-4o-mini}, \textbf{GPT-4o}, \textbf{Claude-3.5-Haiku}, \textbf{Qwen2.5-7B} and \textbf{Qwen3-235B} models. For Step Prompting, models are guided to ``\texttt{think step-by-step}'', explicitly articulating their reasoning process before generating \vis. Models utilizing this approach are denoted by a suffix ``-Step'' (e.g., \textbf{GPT-4o-mini-Step}, \textbf{GPT-4o-Step}, \textbf{Claude-3.5-Haiku-Step}, \textbf{Qwen2.5-7B-Step}, \textbf{Qwen3-235B-Step})

\etitle{Supervised Fine-tuning Method.}
We performed supervised fine-tuning on the Qwen2.5-7B-Instruct model, resulting in a baseline model named \textbf{Qwen2.5-7B-SFT}. This model was trained using the standard SFT approach in the training set, enabling the direct generation of multiple \vis answers for the interpretation of ambiguities.


\etitle{Preference Learning Method.}
We developed an advanced model, referred to as \textbf{\model}, designed to handle ambiguity in \nlvis through step-wise reasoning as detailed in Section~\ref{sec:model}. Following an initial supervised fine-tuning, we constructed a preference dataset from the \data development set specifically for the preference training of \model.

\textbf{Evaluation Metrics.} 
Following prior work~\cite{linenet,ncnet2022,nvBench2021}, we adopt the following metrics:
\textbf{Precision@K (P@K)}: Measures the proportion of valid \vis in the top K output, reflecting the precision of the recommendation.
\textbf{Recall@K (R@K)}: Evaluates the proportion of valid \vis identified, indicating coverage of the golden \vis space.
\textbf{F1@K}: Balances precision and recall, ensuring both high accuracy and comprehensive coverage.
All metrics are reported for $K \in {1, 3, 5}$ to assess performance across varying recommendation set sizes.

\subsection{Experimental Results}

\begin{table*}[t!] 
\centering
\caption{Overall performance comparison between different models on \data. The table presents Recall@K, Precision@K, and F1@K metrics across different model families. Rows 1--12 shows results for models using prompting-based method. The last two rows, Qwen2.5-7B-SFT and \model (ours), are models using supervised fine-tuning method and preference learning method with optimization on \data. \textbf{Bold values} indicate the best performance for each metric. }

\small 
\begin{tabular}{lccccccccc} 
\toprule
\multirow{2}{*}{Model} & \multicolumn{3}{c}{\textbf{Recall@K(\%)}} & \multicolumn{3}{c}{\textbf{Precision@K(\%)}} & \multicolumn{3}{c}{\textbf{F1@K(\%)}} \\
\cline{2-10}         
\\[0.1pt]
& K=1         & K=3         & K=5         & K=1         & K=3        & K=5        & K=1         & K=3        & K=5        \\ 
\midrule
GPT-4o-mini                & 34.72      & 51.92      & 54.65      & 91.88       & 86.86      & 81.76       & 49.31      & 59.73       & 57.60      \\ 
GPT-4o                & 36.56      & 46.35      & 46.79      & 97.07       & 95.83     & \textbf{95.52}       & 51.96      & 58.96       & 59.03      \\
Claude-3.5-Haiku & 36.03	& 67.95	& 67.95 & 95.74	& 93.92	& 93.83	&51.22	&75.63	& 75.56\\
Qwen2.5-7B      &  34.65        & 46.20   &   47.17   &    92.68    &   90.68    &   89.33   &   49.34   &   57.09   &   56.67   \\
Qwen3-235B & 29.37	& 55.59	& 59.39 & 78.83	& 72.70	& 64.84 & 	41.87	& 58.77	& 55.39	\\
\midrule
GPT-4o-mini-Step      & 35.13       & 47.68      & 47.91      & 93.48        & 92.54      & 92.08       & 49.96       & 59.29      & 59.10      \\  
GPT-4o-Step      & 36.30       & 48.92      & 49.21      & 96.94        & 95.47      & 95.08       & 51.72       & 60.78      & 60.66      \\ 
Claude-3.5-Haiku-Step & 35.70	&65.38	&65.52	&94.67	&92.23	&91.97&50.75	&72.84	&72.75\\
Qwen2.5-7B-Step      &  35.20        &  61.86  &   64.08   &   93.61     &  89.26     &   86.23   &   50.05   &   68.56  &   67.76   \\
Qwen3-235B-Step & \textbf{37.49}	&72.83	&75.39 & \textbf{99.60}	& \textbf{95.52}	& 92.21 & \textbf{53.29}	&78.62	&77.78\\
\midrule
Qwen2.5-7B-SFT      & 33.23      & 73.44   & 76.32       & 88.42       & 83.36       & 80.18      & 47.26     & 75.79      & 75.30     \\ 
\midrule
\textbf{\model (ours)}   & 37.30 & \textbf{77.09} & \textbf{79.74} & 99.20  & 94.27 & 91.17 & 53.04     & \textbf{81.50}       & \textbf{80.88}       \\ 
\bottomrule
\end{tabular} 
\label{tab:overall_exp}
\end{table*}

\stitle{Overall Results.}
Table~\ref{tab:overall_exp} presents the comprehensive performance evaluation of different models on \data. 
Our proposed \model achieves state-of-the-art performance across most metrics, significantly outperforming both prompting-based and fine-tuning-based baselines. Specifically, \model obtains the highest F1@3 (81.50\%) and F1@5 (80.88\%), demonstrating its superior ability to handle ambiguity in \nlvis tasks. 
Step-wise reasoning consistently improves performance across most models, although the benefits vary by model architecture. While some models achieve higher recall or precision in specific scenarios, they fail to maintain competitive F1 scores, indicating an imbalance between precision and recall. For GPT-4o, Qwen2.5-7B, and Qwen-3-235B models, the ``-Step'' variants show notable improvements in F1 scores compared to their direct prompting counterparts. This validates our hypothesis that decomposing complex \vis reasoning into explicit steps helps resolve ambiguity more effectively. 
Fine-tuning on \data substantially improves recall at higher $K$ values. Qwen2.5-7B-SFT achieves 75.79\% F1@3 and 75.30\% F1@5, significantly outperforming prompt-based methods of similar model size and those from the same model family, indicating superior coverage of the valid \vis space. However, this approach sacrifices some precision compared to prompting-based methods. 
Finally, our preference-optimized \model achieves the best balance between precision and recall. At $K=1$, it maintains exceptional precision (99.20\%) while improving recall over all baselines. At $K=3$ and $K=5$, it achieves substantial gains in recall without significant precision degradation, demonstrating the effectiveness of step-wise preference optimization for ambiguous \nlvis tasks.

\begin{figure*}[t!]
    \centering
    \includegraphics[width=\textwidth]{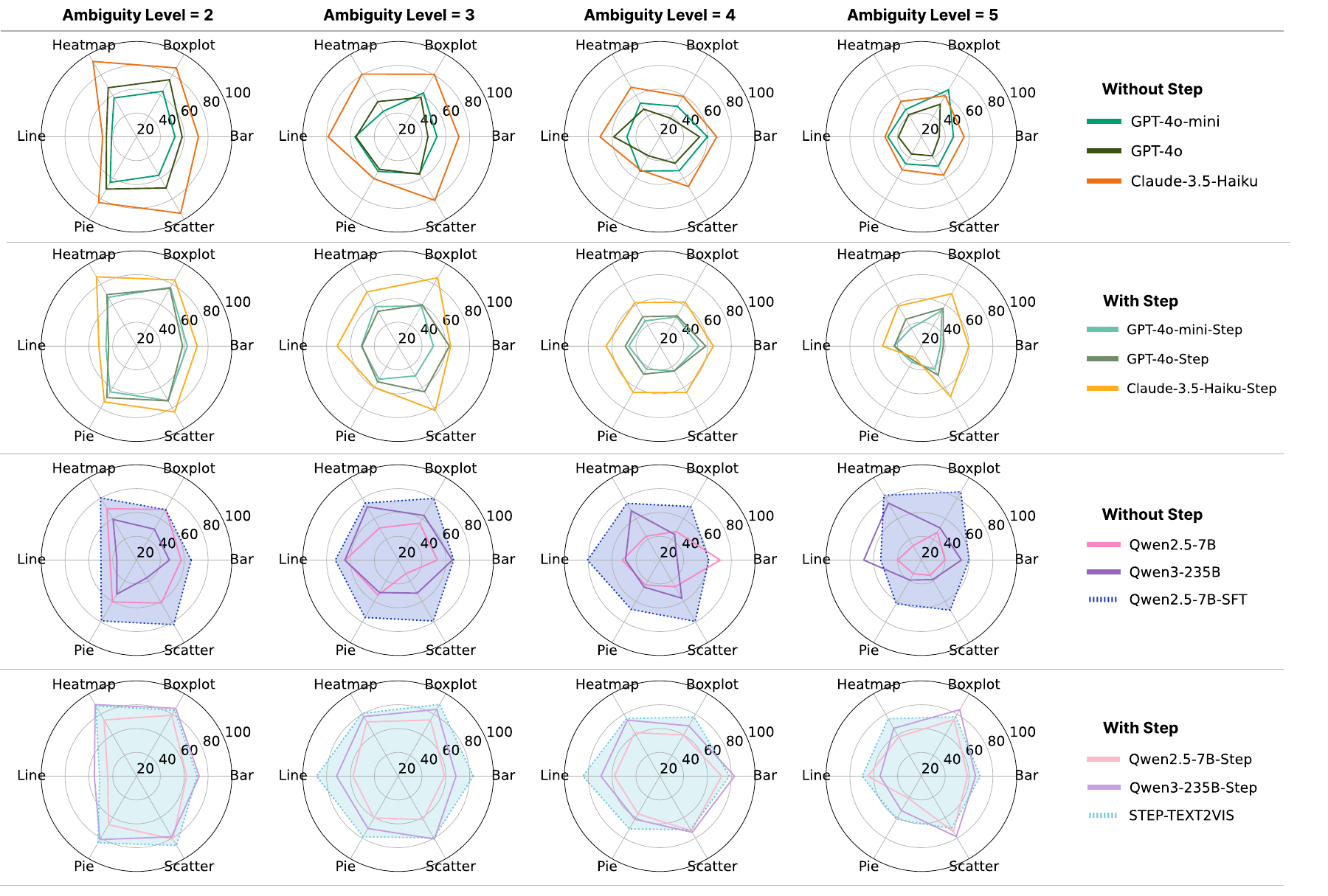}
    \caption{F1 scores across different models and ambiguity levels. The figure is organized as a 4$\times$4 grid where columns represent increasing ambiguity levels, and rows represent different model groups. The first two rows display GPT, Claude model families with prompting-based methods. The last two rows display Qwen model families of prompting-based, supervised and preference learning methods, including our proposed \model in the bottom row. Each radar chart displays F1@5 scores across six chart types, where larger polygons indicate better performance.}
    \label{fig:f1_k}
\end{figure*}

\textbf{Performance Analysis Across Chart Types.}
Figure~\ref{fig:f1_k} presents a radar chart of F1@5 scores for different methods across various chart types and ambiguity levels. 
We have the following observations.

First, \model consistently outperforms other models across most chart types and ambiguity levels. These results demonstrate that the step-wise reasoning approach significantly enhances performance on ambiguous \nlvis tasks.
Second, models with step-wise reasoning (those with ``-Step'' suffix) generally outperform direct prompting models, confirming the effectiveness of breaking complex \vis reasoning into explicit steps.

\begin{itemize}
    \item \textbf{The Impact of Chart Types.} The experimental results reveal that different chart types exhibit varying challenges for models. Boxplot and Scatter charts generally achieve higher F1 scores, indicating they are easier for models to handle. In contrast, Pie charts perform worse at higher ambiguity levels, while Line charts consistently show lower accuracy across all ambiguity levels, with F1 scores only around 40\% to 51\%, even at lower ambiguity levels. These findings suggest that certain chart types pose greater challenges for model interpretation and generation.
    \item \textbf{The Impact of Ambiguity Levels.} The data shows a clear degradation in performance as the ambiguity level increases: At ambiguity level 2, most models maintain relatively high F1 scores (60-80\%). By ambiguity level 5, even the best performing models struggle to maintain the same level of performance.
    For instance, Claude-3.5-Haiku and Claude-3.5-Haiku-Step maintain over 80\% F1 scores on Heatmap, Boxplot, Pie, and Scatter charts at ambiguity level 2, yet decline to below 60\% for these chart types at ambiguity level 5.
    Moreover, Qwen2.5-7B-Step achieving 41.55\% and \model achieving 61\% F1 score for pie charts at this highest ambiguity level. This pattern confirms the inherent challenge of handling highly ambiguous text queries.
    \item \textbf{Step-wise reasoning enhances performance but alters strengths for certain models.} Most prompting-based models exhibit improvements in performance when utilizing step-wise reasoning, while still maintaining their original strengths across chart types and ambiguity levels. However, for Qwen2.5-7B, the introduction of step-wise reasoning leads to notable shifts in its area of expertise. Specifically, Qwen2.5-7B-Step demonstrates significant improvements in Boxplot (74.47\%) and Heatmap (72.59\%) generation at Ambiguity Level 3—an enhancement that was not prominently observed in the base Qwen2.5-7B model. This suggests that step-wise reasoning not only improves overall performance but also reshapes the model's proficiency across different tasks.
\end{itemize}

\stitle{Performance Analysis on Ambiguity Resolution Ability.}
Figure~\ref{fig:Recall_k} illustrates the Recall@5 metric, measuring each model's ability to generate valid \vis from text queries with varying ambiguity levels.
Our \model shows superior recall performance across all ambiguity levels. At ambiguity level 3, it achieves 83.3\% recall, significantly outperforming other models. Further analysis across ambiguity levels reveals the following insights:

\begin{figure}[t!]
    \centering
    \includegraphics[width=\linewidth]{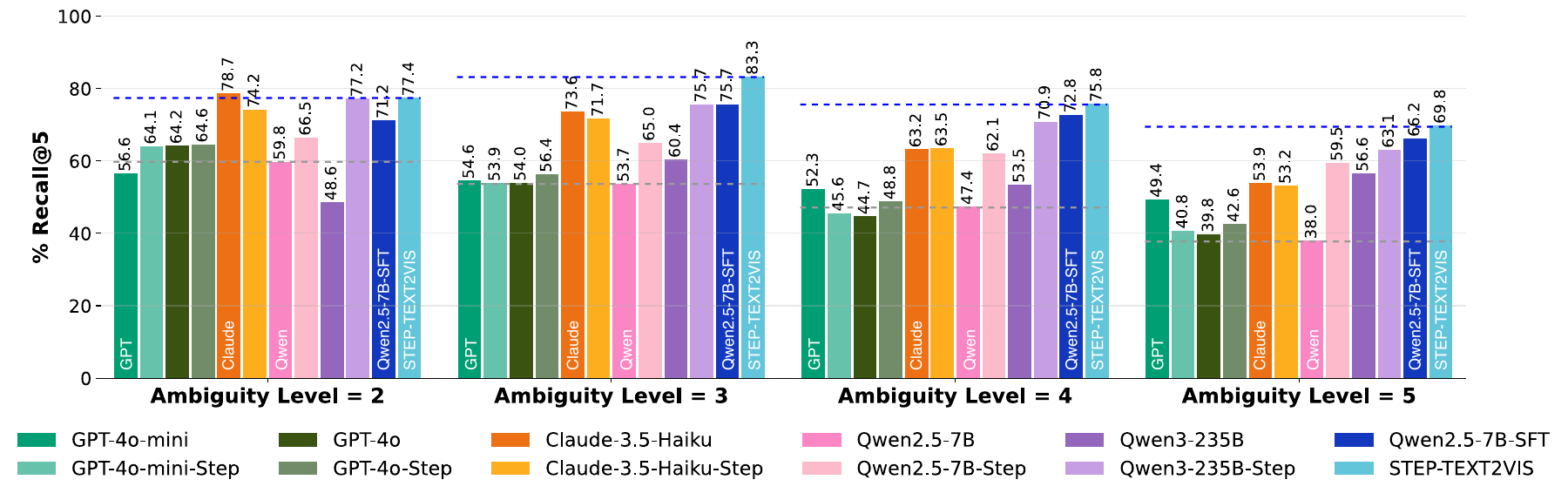}
    \caption{Recall@5 across different models and ambiguity levels. The blue dashed horizontal line indicates the performance of our proposed \model method, while the grey dashed horizontal line represents Qwen2.5-7B-SFT, which serves as the base model for our approach.}
    \label{fig:Recall_k}
\end{figure}

\begin{itemize}
    \item \textbf{Step-wise reasoning significantly enhances performance.} Models implementing step-by-step reasoning methodologies consistently demonstrate superior performance compared to their non-stepwise counterparts. For example, Qwen2.5-7B-Step exhibits markedly improved performance metrics relative to the base Qwen2.5-7B implementation.

    \item \textbf{Inverse correlation between performance and ambiguity.} The experimental results indicate a consistent negative correlation between recall performance and ambiguity level for the majority of models evaluated. This trend confirms the inherently increasing complexity of \vis generation as the ambiguity level intensifies.
    
    \item \textbf{Maximum performance differentiation occurs at intermediate levels of ambiguity.} The performance delta between the evaluated models reaches its maximum when AL equals 3 or 4, suggesting that these intermediate levels provide optimal conditions for discriminating between different model capabilities.
    
    \item \textbf{Fine-tuning methods yield robust performance under increasing ambiguity.} While performance degradation is observed across all models as ambiguity increases, models employing Supervised Fine-Tuning and Preference Learning methodologies maintain superior performance characteristics at elevated ambiguity levels. Notably, the performance differential between \model and alternative approaches expands proportionally with increasing ambiguity.
\end{itemize}

\textbf{Key Implications for Ambiguous \nlvis Systems.}
Our experimental results reveal several important implications for the design of \nlvis systems that can effectively handle ambiguity.
First, the performance improvements achieved through step-wise reasoning highlight the importance of decomposing complex tasks into interpretable steps, similar to how humans reason through ambiguous text queries, rather than relying on direct translation.
Second, the observed performance variations across different chart types and ambiguity levels suggest that future \nlvis systems should adaptively select reasoning strategies based on both the queries' characteristics and the target \vis type. 
Third, the superior performance of \model, particularly at higher ambiguity levels, demonstrates that preference-optimized models can learn to effectively balance precision and recall, maintaining high accuracy while capturing the full range of valid interpretations. 
These findings point toward a paradigm shift in \model development: from single-output systems toward multi-interpretation frameworks that explicitly model and resolve ambiguity through structured reasoning processes.

\section{Related Work}
\label{sec:related}

\textbf{\nlvis benchmarks.}
As the predecessor of \data, \nvbench 1.0~\cite{nvBench2021} is a commonly used \nlvis benchmark that leverages the semantic alignment between queries and Visualization Query Language to construct datasets. Building on this, benchmarks like Dial-NVBench~\cite{dail-nvbench2023} introduce multi-turn dialogues, and VisEval~\cite{VisEval2024} further expand \nlvis evaluation. However, they primarily focus on well-specified queries with a single correct visualization. While datasets like ChartGPT~\cite{ChartGPT2025}, Text2Analysis~\cite{Text2Analysis2023}, QUDA~\cite{QUDA2020}, and NLV~\cite{nlv2021} include ambiguous queries, they lack explicit ambiguity type definitions and comprehensive valid chart sets. In contrast, we introduce \data, the first ambiguity-aware \nlvis benchmark designed to address this gap.

\textbf{LLMs for Data Synthesis.}
LLMs have shown promise in data synthesis across various domains, enhancing data diversity and model generalization~\cite{SynthBio2021,ToxiGen2022,statqa_nips,askchart,DBLP:journals/corr/abs-2412-07673, alpha-sql, elliesql, nl2sql-bugs, deepvis, AdaTest2022, LAMBADA2019, GPT3Mix2021, DailyDialog2022}. This includes the \nlvis domain, where VL2NL~\cite{VL2NL2024} utilizes LLMs to generate descriptions from \vis. Our approach shares a ``reverse engineering'' philosophy with ScienceBenchmark~\cite{ScienceBenchmark2023}, and generates queries from ambiguity-aware \vis tree to capture ambiguity in \nlvis pairs, leveraging the structured nature of \vis to define ambiguity types. Furthermore, we leverage LLMs to generate multi-step reasoning data, following the effectiveness demonstrated in works like Hunter et al.~\cite{LetsVS2023} and Step-DPO~\cite{StepDPO2024}, to improve model reasoning and interpretability in \nlvis tasks.

A more detailed discussion of related work can be found in Section~\ref{apdx:related_work} in the Appendix.
\section{Conclusion}
\label{sec:conclusion}
In this work, we introduced \dataset, the first benchmark designed for evaluating \nlvis systems in scenarios involving ambiguous user queries. \dataset was generated through a controlled ambiguity-injection pipeline, guaranteeing valid and interpretable results while offering step-wise disambiguation reasoning paths. By using \dataset, we offer a robust framework to assess \nlvis systems' ability to handle ambiguities that arise in real-world applications.

We also proposed \model, an LLM-based \nlvis model trained on \dataset, which significantly improves \nlvis performance in ambiguous scenarios by applying step-wise preference optimization. Our experimental results demonstrate that \model outperforms all existing baselines, establishing a new state-of-the-art for handling ambiguity in \nlvis tasks.





\begin{ack}
This paper was supported by
Young Talent Support Project of Guangzhou Association for Science and Technology (QT-2025-001); the NSF of China (62402409); Guangdong Basic and
Applied Basic Research Foundation (2023A1515110545); Guangzhou
Basic and Applied Basic Research Foundation (2025A04J3935);
Guangzhou-HKUST(GZ) Joint Funding Program (2025A03J3714); and
Guangdong Provincial Project (2023CX10X008).
\end{ack}


\renewcommand{\bibsection}{%
  \section*{References}\medskip
}

\begingroup          
\small               
\bibliographystyle{unsrt}   
\bibliography{sample}       
\endgroup


\newpage


\section*{NeurIPS Paper Checklist}

\begin{enumerate}

\item {\bf Claims}
    \item[] Question: Do the main claims made in the abstract and introduction accurately reflect the paper's contributions and scope?
    \item[] Answer: \answerYes{}
    \item[] Justification: The abstract and introduction clearly outline the development of the dataset, the model, and their performance in handling ambiguous Text-to-Visualization tasks, aligning with the contributions detailed in Sections~\ref{sec:pre} to Section~\ref{sec:model}.
    \item[] Guidelines:
    \begin{itemize}
        \item The answer NA means that the abstract and introduction do not include the claims made in the paper.
        \item The abstract and/or introduction should clearly state the claims made, including the contributions made in the paper and important assumptions and limitations. A No or NA answer to this question will not be perceived well by the reviewers. 
        \item The claims made should match theoretical and experimental results, and reflect how much the results can be expected to generalize to other settings. 
        \item It is fine to include aspirational goals as motivation as long as it is clear that these goals are not attained by the paper. 
    \end{itemize}

\item {\bf Limitations}
    \item[] Question: Does the paper discuss the limitations of the work performed by the authors?
    \item[] Answer: \answerYes{}
    \item[] Justification: The paper discusses limitations such as the scope of datasets, computational constraints, and potential biases in the ambiguity-injection pipeline in Section~\ref{apdx:limit}, addressing the constraints of the work.
    \item[] Guidelines:
    \begin{itemize}
        \item The answer NA means that the paper has no limitation while the answer No means that the paper has limitations, but those are not discussed in the paper. 
        \item The authors are encouraged to create a separate "Limitations" section in their paper.
        \item The paper should point out any strong assumptions and how robust the results are to violations of these assumptions (e.g., independence assumptions, noiseless settings, model well-specification, asymptotic approximations only holding locally). The authors should reflect on how these assumptions might be violated in practice and what the implications would be.
        \item The authors should reflect on the scope of the claims made, e.g., if the approach was only tested on a few datasets or with a few runs. In general, empirical results often depend on implicit assumptions, which should be articulated.
        \item The authors should reflect on the factors that influence the performance of the approach. For example, a facial recognition algorithm may perform poorly when image resolution is low or images are taken in low lighting. Or a speech-to-text system might not be used reliably to provide closed captions for online lectures because it fails to handle technical jargon.
        \item The authors should discuss the computational efficiency of the proposed algorithms and how they scale with dataset size.
        \item If applicable, the authors should discuss possible limitations of their approach to address problems of privacy and fairness.
        \item While the authors might fear that complete honesty about limitations might be used by reviewers as grounds for rejection, a worse outcome might be that reviewers discover limitations that aren't acknowledged in the paper. The authors should use their best judgment and recognize that individual actions in favor of transparency play an important role in developing norms that preserve the integrity of the community. Reviewers will be specifically instructed to not penalize honesty concerning limitations.
    \end{itemize}

\item {\bf Theory assumptions and proofs}
    \item[] Question: For each theoretical result, does the paper provide the full set of assumptions and a complete (and correct) proof?
    \item[] Answer: \answerNA{}
    \item[] Justification: The paper does not present theoretical results or proofs, focusing instead on empirical evaluations and dataset synthesis.
    \item[] Guidelines:
    \begin{itemize}
        \item The answer NA means that the paper does not include theoretical results. 
        \item All the theorems, formulas, and proofs in the paper should be numbered and cross-referenced.
        \item All assumptions should be clearly stated or referenced in the statement of any theorems.
        \item The proofs can either appear in the main paper or the supplemental material, but if they appear in the supplemental material, the authors are encouraged to provide a short proof sketch to provide intuition. 
        \item Inversely, any informal proof provided in the core of the paper should be complemented by formal proofs provided in appendix or supplemental material.
        \item Theorems and Lemmas that the proof relies upon should be properly referenced. 
    \end{itemize}

\item {\bf Experimental result reproducibility}
    \item[] Question: Does the paper fully disclose all the information needed to reproduce the main experimental results of the paper to the extent that it affects the main claims and/or conclusions of the paper (regardless of whether the code and data are provided or not)?
    \item[] Answer: \answerYes{}
    \item[] Justification: Section~\ref{sec:exp} details the experimental setup, including dataset splits, models, and evaluation metrics, while Section~\ref{apdx:exp} in appendix provide further details on the synthesis pipeline and training setup, enabling reproduction of the results.
    \item[] Guidelines:
    \begin{itemize}
        \item The answer NA means that the paper does not include experiments.
        \item If the paper includes experiments, a No answer to this question will not be perceived well by the reviewers: Making the paper reproducible is important, regardless of whether the code and data are provided or not.
        \item If the contribution is a dataset and/or model, the authors should describe the steps taken to make their results reproducible or verifiable. 
        \item Depending on the contribution, reproducibility can be accomplished in various ways. For example, if the contribution is a novel architecture, describing the architecture fully might suffice, or if the contribution is a specific model and empirical evaluation, it may be necessary to either make it possible for others to replicate the model with the same dataset, or provide access to the model. In general, releasing code and data is often one good way to accomplish this, but reproducibility can also be provided via detailed instructions for how to replicate the results, access to a hosted model (e.g., in the case of a large language model), releasing of a model checkpoint, or other means that are appropriate to the research performed.
        \item While NeurIPS does not require releasing code, the conference does require all submissions to provide some reasonable avenue for reproducibility, which may depend on the nature of the contribution. For example
        \begin{enumerate}
            \item If the contribution is primarily a new algorithm, the paper should make it clear how to reproduce that algorithm.
            \item If the contribution is primarily a new model architecture, the paper should describe the architecture clearly and fully.
            \item If the contribution is a new model (e.g., a large language model), then there should either be a way to access this model for reproducing the results or a way to reproduce the model (e.g., with an open-source dataset or instructions for how to construct the dataset).
            \item We recognize that reproducibility may be tricky in some cases, in which case authors are welcome to describe the particular way they provide for reproducibility. In the case of closed-source models, it may be that access to the model is limited in some way (e.g., to registered users), but it should be possible for other researchers to have some path to reproducing or verifying the results.
        \end{enumerate}
    \end{itemize}

\item {\bf Open access to data and code}
    \item[] Question: Does the paper provide open access to the data and code, with sufficient instructions to faithfully reproduce the main experimental results, as described in supplemental material?
    \item[] Answer: \answerYes{}
    \item[] Justification: The paper states in the abstract that source code and data are available at \url{https://nvbench2.github.io/}, providing open access to reproduce the experimental results.
    \item[] Guidelines:
    \begin{itemize}
        \item The answer NA means that paper does not include experiments requiring code.
        \item Please see the NeurIPS code and data submission guidelines (\url{https://nips.cc/public/guides/CodeSubmissionPolicy}) for more details.
        \item While we encourage the release of code and data, we understand that this might not be possible, so “No” is an acceptable answer. Papers cannot be rejected simply for not including code, unless this is central to the contribution (e.g., for a new open-source benchmark).
        \item The instructions should contain the exact command and environment needed to run to reproduce the results. See the NeurIPS code and data submission guidelines (\url{https://nips.cc/public/guides/CodeSubmissionPolicy}) for more details.
        \item The authors should provide instructions on data access and preparation, including how to access the raw data, preprocessed data, intermediate data, and generated data, etc.
        \item The authors should provide scripts to reproduce all experimental results for the new proposed method and baselines. If only a subset of experiments are reproducible, they should state which ones are omitted from the script and why.
        \item At submission time, to preserve anonymity, the authors should release anonymized versions (if applicable).
        \item Providing as much information as possible in supplemental material (appended to the paper) is recommended, but including URLs to data and code is permitted.
    \end{itemize}

\item {\bf Experimental setting/details}
    \item[] Question: Does the paper specify all the training and test details (e.g., data splits, hyperparameters, how they were chosen, type of optimizer, etc.) necessary to understand the results?
    \item[] Answer: \answerYes{}
    \item[] Justification: Section~\ref{sec:exp} describes the dataset splits, models, and evaluation metrics, while Appendix~\ref{apdx:exp} provides additional details on training setup and prompt templates, sufficient to understand the results.
    \item[] Guidelines:
    \begin{itemize}
        \item The answer NA means that the paper does not include experiments.
        \item The experimental setting should be presented in the core of the paper to a level of detail that is necessary to appreciate the results and make sense of them.
        \item The full details can be provided either with the code, in appendix, or as supplemental material.
    \end{itemize}

\item {\bf Experiment statistical significance}
    \item[] Question: Does the paper report error bars suitably and correctly defined or other appropriate information about the statistical significance of the experiments?
    \item[] Answer: \answerNA{}
    \item[] Justification: The paper reports performance metrics (\eg Precision and Recall) in Section\ref{apdx:exp}, with detailed tables and figures. but does not provide error bars, confidence intervals, or statistical significance tests for the experimental results.
    \item[] Guidelines:
    \begin{itemize}
        \item The answer NA means that the paper does not include experiments.
        \item The authors should answer "Yes" if the results are accompanied by error bars, confidence intervals, or statistical significance tests, at least for the experiments that support the main claims of the paper.
        \item The factors of variability that the error bars are capturing should be clearly stated (for example, train/test split, initialization, random drawing of some parameter, or overall run with given experimental conditions).
        \item The method for calculating the error bars should be explained (closed form formula, call to a library function, bootstrap, etc.)
        \item The assumptions made should be given (e.g., Normally distributed errors).
        \item It should be clear whether the error bar is the standard deviation or the standard error of the mean.
        \item It is OK to report 1-sigma error bars, but one should state it. The authors should preferably report a 2-sigma error bar than state that they have a 96\% CI, if the hypothesis of Normality of errors is not verified.
        \item For asymmetric distributions, the authors should be careful not to show in tables or figures symmetric error bars that would yield results that are out of range (e.g. negative error rates).
        \item If error bars are reported in tables or plots, The authors should explain in the text how they were calculated and reference the corresponding figures or tables in the text.
    \end{itemize}

\item {\bf Experiments compute resources}
    \item[] Question: For each experiment, does the paper provide sufficient information on the computer resources (type of compute workers, memory, time of execution) needed to reproduce the experiments?
    \item[] Answer: \answerYes{}
    \item[] Justification: The paper specifies the compute resources (e.g., CPU/GPU type, memory, execution time) used for the experiments in the appendix~\ref{apdx:exp}, providing sufficient details for reproducibility.
    \item[] Guidelines:
    \begin{itemize}
        \item The answer NA means that the paper does not include experiments.
        \item The paper should indicate the type of compute workers CPU or GPU, internal cluster, or cloud provider, including relevant memory and storage.
        \item The paper should provide the amount of compute required for each of the individual experimental runs as well as estimate the total compute. 
        \item The paper should disclose whether the full research project required more compute than the experiments reported in the paper (e.g., preliminary or failed experiments that didn't make it into the paper). 
    \end{itemize}

\item {\bf Code of ethics}
    \item[] Question: Does the research conducted in the paper conform, in every respect, with the NeurIPS Code of Ethics \url{https://neurips.cc/public/EthicsGuidelines}?
    \item[] Answer: \answerYes{}
    \item[] Justification: The research adheres to NeurIPS Code of Ethics.
    \item[] Guidelines:
    \begin{itemize}
        \item The answer NA means that the authors have not reviewed the NeurIPS Code of Ethics.
        \item If the authors answer No, they should explain the special circumstances that require a deviation from the Code of Ethics.
        \item The authors should make sure to preserve anonymity (e.g., if there is a special consideration due to laws or regulations in their jurisdiction).
    \end{itemize}

\item {\bf Broader impacts}
    \item[] Question: Does the paper discuss both potential positive societal impacts and negative societal impacts of the work performed?
    \item[] Answer: \answerYes{}
    \item[] Justification: The paper discusses potential positive and negative societal impacts, such as the benefits of improved visualization tools and risks of misuse for disinformation or fairness issues, in Appendix~\ref{apdx:ethic}.
    \item[] Guidelines:
    \begin{itemize}
        \item The answer NA means that there is no societal impact of the work performed.
        \item If the authors answer NA or No, they should explain why their work has no societal impact or why the paper does not address societal impact.
        \item Examples of negative societal impacts include potential malicious or unintended uses (e.g., disinformation, generating fake profiles, surveillance), fairness considerations (e.g., deployment of technologies that could make decisions that unfairly impact specific groups), privacy considerations, and security considerations.
        \item The conference expects that many papers will be foundational research and not tied to particular applications, let alone deployments. However, if there is a direct path to any negative applications, the authors should point it out. For example, it is legitimate to point out that an improvement in the quality of generative models could be used to generate deepfakes for disinformation. On the other hand, it is not needed to point out that a generic algorithm for optimizing neural networks could enable people to train models that generate Deepfakes faster.
        \item The authors should consider possible harms that could arise when the technology is being used as intended and functioning correctly, harms that could arise when the technology is being used as intended but gives incorrect results, and harms following from (intentional or unintentional) misuse of the technology.
        \item If there are negative societal impacts, the authors could also discuss possible mitigation strategies (e.g., gated release of models, providing defenses in addition to attacks, mechanisms for monitoring misuse, mechanisms to monitor how a system learns from feedback over time, improving the efficiency and accessibility of ML).
    \end{itemize}

\item {\bf Safeguards}
    \item[] Question: Does the paper describe safeguards that have been put in place for responsible release of data or models that have a high risk for misuse (e.g., pretrained language models, image generators, or scraped datasets)?
    \item[] Answer: \answerNA{}
    \item[] Justification: The paper’s dataset and models do not pose high risks for misuse, as they focus on Text-to-Visualization tasks without sensitive data or dual-use potential.
    \item[] Guidelines:
    \begin{itemize}
        \item The answer NA means that the paper poses no such risks.
        \item Released models that have a high risk for misuse or dual-use should be released with necessary safeguards to allow for controlled use of the model, for example by requiring that users adhere to usage guidelines or restrictions to access the model or implementing safety filters. 
        \item Datasets that have been scraped from the Internet could pose safety risks. The authors should describe how they avoided releasing unsafe images.
        \item We recognize that providing effective safeguards is challenging, and many papers do not require this, but we encourage authors to take this into account and make a best faith effort.
    \end{itemize}

\item {\bf Licenses for existing assets}
    \item[] Question: Are the creators or original owners of assets (e.g., code, data, models), used in the paper, properly credited and are the license and terms of use explicitly mentioned and properly respected?
    \item[] Answer: \answerYes{}
    \item[] Justification: The paper credits previous datasets (e.g., nvBench and BIRD) and implies adherence to licensing.
    \item[] Guidelines:
    \begin{itemize}
        \item The answer NA means that the paper does not use existing assets.
        \item The authors should cite the original paper that produced the code package or dataset.
        \item The authors should state which version of the asset is used and, if possible, include a URL.
        \item The name of the license (e.g., CC-BY 4.0) should be included for each asset.
        \item For scraped data from a particular source (e.g., website), the copyright and terms of service of that source should be provided.
        \item If assets are released, the license, copyright information, and terms of use in the package should be provided. For popular datasets, \url{paperswithcode.com/datasets} has curated licenses for some datasets. Their licensing guide can help determine the license of a dataset.
        \item For existing datasets that are re-packaged, both the original license and the license of the derived asset (if it has changed) should be provided.
        \item If this information is not available online, the authors are encouraged to reach out to the asset's creators.
    \end{itemize}

\item {\bf New assets}
    \item[] Question: Are new assets introduced in the paper well documented and is the documentation provided alongside the assets?
    \item[] Answer: \answerYes{}
    \item[] Justification: The dataset and model are documented in Sections~\ref{sec:pre} and Sections~\ref{sec:model}, with additional details in Appendix~\ref{apdx:dataset}, and the open access URL (\url{https://nvbench2.github.io/}) likely includes further documentation.
    \item[] Guidelines:
    \begin{itemize}
        \item The answer NA means that the paper does not release new assets.
        \item Researchers should communicate the details of the dataset/code/model as part of their submissions via structured templates. This includes details about training, license, limitations, etc. 
        \item The paper should discuss whether and how consent was obtained from people whose asset is used.
        \item At submission time, remember to anonymize your assets (if applicable). You can either create an anonymized URL or include an anonymized zip file.
    \end{itemize}

\item {\bf Crowdsourcing and research with human subjects}
    \item[] Question: For crowdsourcing experiments and research with human subjects, does the paper include the full text of instructions given to participants and screenshots, if applicable, as well as details about compensation (if any)? 
    \item[] Answer: \answerNA{}
    \item[] Justification: The paper does not involve crowdsourcing or human subjects, as the experiments focus on automated data synthesis and model training.
    \item[] Guidelines:
    \begin{itemize}
        \item The answer NA means that the paper does not involve crowdsourcing nor research with human subjects.
        \item Including this information in the supplemental material is fine, but if the main contribution of the paper involves human subjects, then as much detail as possible should be included in the main paper. 
        \item According to the NeurIPS Code of Ethics, workers involved in data collection, curation, or other labor should be paid at least the minimum wage in the country of the data collector. 
    \end{itemize}

\item {\bf Institutional review board (IRB) approvals or equivalent for research with human subjects}
    \item[] Question: Does the paper describe potential risks incurred by study participants, whether such risks were disclosed to the subjects, and whether Institutional Review Board (IRB) approvals (or an equivalent approval/review based on the requirements of your country or institution) were obtained?
    \item[] Answer: \answerNA{}
    \item[] Justification: The paper does not involve human subjects, focusing on computational experiments and dataset synthesis, thus requiring no IRB approval.
    \item[] Guidelines:
    \begin{itemize}
        \item The answer NA means that the paper does not involve crowdsourcing nor research with human subjects.
        \item Depending on the country in which research is conducted, IRB approval (or equivalent) may be required for any human subjects research. If you obtained IRB approval, you should clearly state this in the paper. 
        \item We recognize that the procedures for this may vary significantly between institutions and locations, and we expect authors to adhere to the NeurIPS Code of Ethics and the guidelines for their institution. 
        \item For initial submissions, do not include any information that would break anonymity (if applicable), such as the institution conducting the review.
    \end{itemize}

\item {\bf Declaration of LLM usage}
    \item[] Question: Does the paper describe the usage of LLMs if it is an important, original, or non-standard component of the core methods in this research? Note that if the LLM is used only for writing, editing, or formatting purposes and does not impact the core methodology, scientific rigorousness, or originality of the research, declaration is not required.
    \item[] Answer: \answerYes{}
    \item[] Justification: The paper details the use of LLMs in the ambiguity-injection pipeline (Section~\ref{sec:synthesizer}) and experiments (Section~\ref{sec:exp}), which are central to the core methodology.
    \item[] Guidelines:
    \begin{itemize}
        \item The answer NA means that the core method development in this research does not involve LLMs as any important, original, or non-standard components.
        \item Please refer to our LLM policy (\url{https://neurips.cc/Conferences/2025/LLM}) for what should or should not be described.
    \end{itemize}

\end{enumerate}

\newpage
\appendix

\section*{Technical Appendices and Supplementary Material}
\begin{enumerate}[label=-]
    \item Section~\ref{apdx:related_work}: Detailed Related Work 
    \item Section~\ref{apdx:synthetic}: More Details of Synthetic Pipeline
    \item Section~\ref{apdx:dataset}: More Details of \data
    \item Section~\ref{apdx:exp}: More Details of Experimental Setups
    \item Section~\ref{apdx:error}: More Details of Error Analysis
    \item Section~\ref{apdx:limit}: Limitations
    \item Section~\ref{apdx:ethic}: Ethic Statement
\end{enumerate}

\section{Detailed Related Work}
\label{apdx:related_work}

\subsection{\nlvis Benchmarks}
\nlvis benchmarks play a crucial role in evaluating the performance of \nlvis systems~\cite{DBLP:journals/corr/abs-2404-18144}.  
As the predecessor of \data, \nvbench~1.0~\cite{nvBench2021} is a commonly used \nlvis benchmark, constructs datasets by leveraging the semantic alignment between \sql and visualization query language, which is a SQL-like specification that defines the visualization structure and details the data transformation processes.
It employs template-based structures to systematically translate \vql into NL. This structured approach facilitates end-to-end model training by enhancing the clarity of both inputs and outputs~\cite{ncnet2022,Song,Wu2022b,Author2023,Tian2023,Li2024b}.
Building on nvBench~1.0~\cite{nvBench2021}, Dial-NVBench~\cite{dail-nvbench2023} introduces multi-turn dialogues, allowing models to capture user intent through iterative interactions. VisEval~\cite{VisEval2024} further refines \nvbench by filtering out ambiguous, irrational, duplicated, and incorrect queries using a three-step selection process (rule-based, LLM-based, and human-based), and offers an automated evaluation framework covering validity, legality, and readability.
However, all three benchmarks ~\cite{nvBench2021, dail-nvbench2023, VisEval2024} remain focused on well-specified queries that map directly to a single correct visualization, without explicitly addressing ambiguity in user intent.

To explore ambiguous and under-specified query formulations, ChartGPT~\cite{ChartGPT2025} extends \nvbench by prompting GPT-3 to generate more abstract and natural utterances compared to the original ones.
Similarly, while some other \nlvis datasets include ambiguous queries~\cite{Text2Analysis2023,QUDA2020,nlv2021},
they do not explicitly define ambiguity types and provide a complete set of valid chart results.
Beyond the realm of \nlvis, ambiguity has also been explored in \nlsql benchmarks, where studies have considered data selection and computation ambiguity~\cite{ambiQT2023,AMBROSIA2024}, but they do not address ambiguity in the visualization space. While some \nlvis systems have attempted to address ambiguity by detecting it~\cite{DataTone2015,Eviza2016} or inferring underspecified queries~\cite{Setlur2019}, they lack a benchmark for systematic evaluation. 

To fill this gap, we propose \data, the first ambiguity-aware \nlvis benchmark, which provides ambiguous user queries and supports one-to-many mappings with multiple valid visualizations. By doing so, it enables a more comprehensive evaluation of \nlvis systems in real-world scenarios.

\subsection{LLMs for Data Synthesis}
Recently, the use of LLMs for data synthesis or data augmentation has become increasingly prevalent. Many studies leverage LLM-generated data for training models~\cite{SynthBio2021,ToxiGen2022,statqa_nips,askchart,DBLP:journals/corr/abs-2412-07673, alpha-sql, naturallanguagerewriter}, as well as for evaluating the performance of other trained models~\cite{AdaTest2022}. In the NLP domain, researchers have utilized LLMs to generate synthetic data for tasks like text classification~\cite{LAMBADA2019,GPT3Mix2021,DailyDialog2022}. These works showcase that LLM-generated data can enhance data diversity, thereby improving model generalization and robustness. 
Building on this, VL2NL~\cite{VL2NL2024} extends LLMs to \nlvis domain, generating natural language descriptions (\eg L1 and L2 captions, and user commands) from Vega-Lite specifications. 
%
Similarly, the application of LLMs for tabular data or database-related tasks has gained attraction. Common approaches for generating \nlsql or table question answering datasets often involve generating \nlq queries first, followed by \sql generation~\cite{ambiQT2023,AMBROSIA2024}.
ScienceBenchmark~\cite{ScienceBenchmark2023} takes a reverse approach by starting with seed \sql queries, then generating new \sql queries from the domain schema, and translating them into natural language queries using fine-tuned LLMs. We follow this reverse construction philosophy in developing \data. Specifically, we begin by extracting \vql from seed charts and then use LLMs to reverse engineer the corresponding text descriptions. The advantage of this approach is that \vql clearly defines each step and the ambiguity types involved, allowing us to better capture one-to-many (\nlq, \vis) pairs.

By leveraging LLMs to generate multi-step reasoning data, the performance of models on long-chain and complex reasoning tasks can be further improved. As demonstrated by Hunter et al.~\cite{LetsVS2023}, process supervision via multi-step reasoning significantly enhances model reliability on tasks such as mathematical problem-solving. Similarly, Step-DPO~\cite{StepDPO2024} shows that generating step-wise reasoning data enables models to better capture intermediate steps, resulting in improved accuracy. Following this approach, we also generate multi-step reasoning data for tasks in the \nlvis domain, where each step of the reasoning process is explicitly defined, contributing to more accurate and interpretable model predictions.
\section{More Details of Synthetic Pipeline}
\label{apdx:synthetic}


\subsection{Step 1: Ambiguity-aware VIS Tree Synthesis}  
\label{sub:step1}  

\begin{figure}[t!]
  \centering
  \includegraphics[width=\columnwidth]{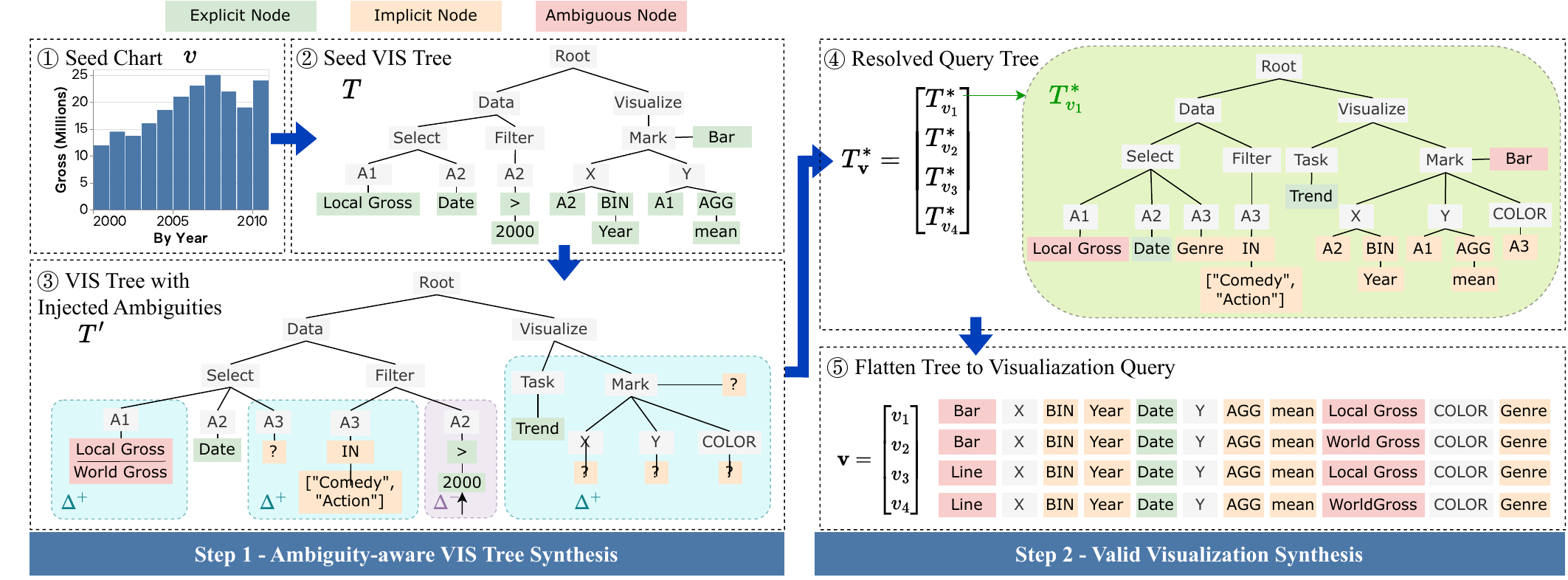}
  \vspace{-2em}
	\caption{Injecting ambiguities into a seed visualization.}
	\label{fig:ambiguity_tree}
\end{figure}

Our ambiguity-aware visualization tree synthesis forms the foundation for synthesizing ambiguous \nlvis data. As shown in Figure~\ref{fig:ambiguity_tree}, this process injects ambiguities into a seed visualization.

\stitle{Transforming the Seed Visualization into a Tree Abstraction.} 
Given a data table $D$ and a seed visualization $v$ (\eg (Figure~\ref{fig:ambiguity_tree}-\ding{172}), we first convert the $v$—along with its underlying query—into an Abstract Syntax Tree (AST), which we refer to as the seed visualization tree $T$ (\eg Figure~\ref{fig:ambiguity_tree}-\ding{173}).
The grammar of AST is based on the predecessor work, nvBench~\cite{nvBench2021}. 
This tree explicitly encodes all design decisions made in creating $v$ and is formally defined as:

\begin{equation}  
v \mapsto T = \{\mathbf{A} \mid \mathbf{A} = [a_1, a_2, \ldots, a_t]\}  
\end{equation}  

Here, each node $a_i$ represents a construction action for a visualization component as a tuple $(\tau, \mathit{op}, \mathit{params})$, where:

\begin{itemize}  
    \item $\tau \in \{\mathtt{explicit}, \mathtt{ambiguous}, \mathtt{implicit}\}$ denotes the ambiguity type of the action node;
    
    \item $\mathit{op}$ specifies the operation (\eg data selection, chart type selection, channel mapping, data transformation selection, etc.);
    
    \item $\mathit{params}$ contains the specific parameters for the operations.
\end{itemize}

\stitle{Controlled Ambiguity Injection.}

We then transform $T$ into an ambiguity-aware tree $T'$ through three operations:  

\begin{itemize}
\item \textsf{\fcolorbox{pink}{pink}{{\small Injecting ambiguous nodes}}}: We add nodes that represent components with multiple valid interpretations. For example, replacing ``Local Gross'' with an ambiguous choice between ``Local Gross'' and ``World Gross''.

\item \textsf{\fcolorbox{lightyellow}{lightyellow}{{\small Adding implicit nodes}}}: We include nodes for components not explicitly specified but required for visualization completion. For example, adding a node for the ``\textsc{color}'' encoding channel.

\item \textsf{\fcolorbox{lightgreen}{lightgreen}{{\small Modifying explicit nodes}}}: We adjust certain explicit nodes to account for potential ambiguities. For example, changing a ``Mark'' node initially set as ``Bar'' into an ambiguous choice among various mark types or requiring inference from analytic tasks.
\end{itemize}

By applying these steps, the resulting ambiguity-aware tree $T'$ captures the full range of possible interpretations for the seed visualization. For example, as shown in Figure~\ref{fig:ambiguity_tree}-\ding{174}, this tree contains some new nodes such as:

\noindent
\textsf{\small \fcolorbox{pink}{pink}{{A1}}: $\left(\text{ambiguous}, \text{data\_column}, \{\text{field:} [\text{Local\_Gross}, \text{World\_Gross}]\}\right)$}

\noindent
\textsf{\small \fcolorbox{lightgreen}{lightgreen}{{A2}}: $\left(\text{explicit}, \text{task}, \{\text{value:} [\text{Trend}]\}\right)$}

\noindent
\textsf{\small\fcolorbox{lightyellow}{lightyellow}{{A3}}: $\left(\text{implicit}, \text{data\_value}, \{\text{value:} [\text{Comedy}, \text{Action}]\}\right)$}

\stitle{Ambiguity Metadata Generation for Ambiguity Injection.}  

\begin{figure}[t!]  
  \centering  
  \includegraphics[width=0.7\columnwidth]{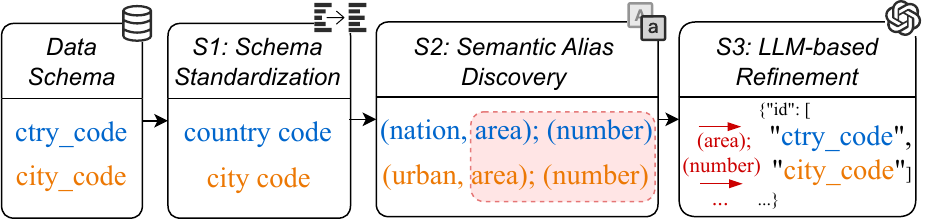}  
  \caption{Ambiguity metadata generation workflow.}  
  \label{fig:ambiguityGeneration}  
   \vspace{-1em}
\end{figure}

To enable precise ambiguity injection in data tables, we propose a systematic metadata generation process that integrates structured knowledge bases with large language models (LLMs). This process identifies and categorizes potential semantic ambiguities in table schemas, producing metadata that guides the construction of ambiguity-aware visualization trees. Each node in the visualization tree \( T' \) is labeled as ambiguous, implicit, or explicit based on the metadata, ensuring visualizations reflect multiple valid query interpretations. The process comprises three key stages: schema standardization, semantic alias discovery, and LLM-based refinement. 

\etitle{Stage 1: Schema Standardization:} The first step involves standardizing the original data schema by refining or expanding column names. Abbreviated or domain-specific terms are transformed into more descriptive, conventional labels. For example, a column labeled {\tt ctry\_code} is standardized to {\tt country code}. Such standardization forms a clearer basis for subsequent ambiguity analysis.

\etitle{Stage 2: Semantic Alias Discovery:}
After standardizing the schema, we leverage ConceptNet~\cite{conceptNet} to identify potential semantic aliases for each column name. ConceptNet's multilingual knowledge graph provides synonyms, hypernyms, and other semantically related terms, helping detect conceptual overlaps. We flag pairs of columns with similar meanings or concept overlap as potential sources of ambiguity. For example, {\tt country code} and {\tt city code} may both have meanings related to {\tt area number}, introducing possible confusion in user queries.

\etitle{Stage 3: LLM-Based Refinement:}
We refine the flagged ambiguous column pairs using GPT-4o-mini with a chain-of-thought (CoT) prompting strategy. The model analyzes the original column names, their standardized forms (Stage 1), and the ConceptNet-derived aliases and ambiguity flags (Stage 2). It then generates a final, validated set of ambiguous pairs, which is formatted into a JSON metadata file. For example, as shown in Figure~\ref{fig:ambiguityGeneration}, one of the identified ambiguous pairs is {\tt ctry\_code} and {\tt city\_code} due to their similar word aliases.
This process supports ambiguity-aware visualization generation and step-wise reasoning.

By combining these stages, we generate the necessary metadata to guide the construction of ambiguity-aware visualization trees, ensuring that each node is accurately marked as explicit, ambiguous, or implicit, thus enabling the synthesis of visualizations that reflect multiple valid interpretations of the query.

\subsection{Step 2: Valid Visualization Synthesis}
\label{sub:step2}

Once we have an ambiguity-aware visualization tree $T'$, the next stage is to generate a set of valid visualizations $\mathbf{v}=\{v_1, v_2, \ldots, v_k\}$.
Each visualization $v_i$ represents one possible resolution of the ambiguities present in $T'$ (see Figure~\ref{fig:ambiguity_tree}-\ding{174}). 
In this step, we define a resolution function $\mathcal{R}$ to systematically clarifies ambiguous and implicit nodes, transforming $T'$ into a set of resolved trees $\{T^*_{v_1}, \ldots, T^*_{v_k}\}$ (see Figure~\ref{fig:ambiguity_tree}-\ding{175}).  Each resolved tree $T^*_{v_i}$ is then ``flattened'' into a concrete visualization query $v_i$ (see Figure~\ref{fig:ambiguity_tree}-\ding{176}).

\stitle{Task Description.} Recap that a partially ambiguous visualization tree $T'$ may contain: 

\begin{itemize}
\item \textit{Ambiguous nodes}: Multiple valid interpretations (\eg which column to use for ``gross'').
\item \textit{Implicit nodes}: Necessary but unspecified details (\eg binning a date field by year).
\item \textit{Explicit nodes}: Directly specified components (\eg ``bar'' mark).
\end{itemize}

To produce valid visualizations, these ambiguous and implicit nodes must be resolved in a manner consistent with established visualization grammar rules (\eg requiring temporal fields to be binned). Formally, we define: 

\begin{equation}
\mathcal{R}(T') \rightarrow \{T^*_{v_1}, T^*_{v_2}, \ldots, T^*_{v_k}\}
\end{equation}

where each $T^*_{v_i}$ is a resolved tree that has no remaining ambiguity or unspecified details. The flattening process then converts each $T^*_{v_i}$ into a finalized visualization specification $v_i$. This yields the complete set of valid visualizations: $\mathbf{v}=\{v_1, v_2, \ldots, v_k\}$.

In the following sections, we describe how an Answer Set Programming (\textbf{ASP}) solver~\cite{gebser2019multi} is used to implement the resolution function 
$\mathcal{R}$ while ensuring that each resolved visualization adheres to the necessary grammar constraints.

\stitle{ASP Solver Objective.}
ASP is a declarative constraint programming paradigm well-suited for knowledge representation and reasoning~\cite {gebser2019multi,moritz2018draco,yang2023draco2}. 
Encoding the ambiguity resolution process and grammar rules as logical constraints has the following benefits:

\begin{itemize}
\item \textit{Completeness}: The solver can enumerate all stable models (\ie all possible ways to resolve ambiguous or implicit nodes) that satisfy the visualization grammar.
\item \textit{Correctness}: Only solutions that meet mandatory constraints (\eg ``temporal fields must be binned'') are considered valid.
\item \textit{Diversity}: Each output corresponds to a distinct interpretation of the query, ensuring coverage of all plausible visualizations.
\end{itemize}

The number of resulting visualizations, $k=|\textbf{v}|$, represents the \textbf{ambiguity level}—how many distinct interpretations the solver deems valid for the given $T'$. 
After obtaining these solutions, we can filter or select a subset based on a target ambiguity level $k$, ensuring that each retained visualization differs from the others.

\stitle{ASP Syntax Overview.} 
ASP is built on a logical foundation with several key syntactic constructs~\cite{gebser2019multi}. The fundamental unit in ASP is a rule of the form: \asprule{~Head :- Body.}, which states that the head is true if all literals in the body are satisfied. 
For example, the rule: \asprule{~light\_on :- power\_available, switch\_flipped.} expresses that the light will be on if both power is available and the switch is flipped.

Some special cases include: 
\begin{itemize}
    \item \textit{Facts}: Rules without a body represent unconditional truths. For example, \asprule{~power\_available.} asserts that power is available.
    \item \textit{Integrity Constraints}: Rules without a head prohibit certain combinations of conditions. For example, the constraint: \asprule{~:- not power\_available, light\_on.} ensures that the light cannot be on when power is not available.
\end{itemize}

An ASP program consists of a collection of rules, facts, and constraints that collectively define a search space. The ASP solver then computes all stable models (\ie answer sets) that satisfy these conditions. Each stable model represents a valid system state or, in our context, a valid resolution of the ambiguous visualization tree.

For example, consider a simple lighting system modeled with:

\begin{itemize}
\item \textit{Rule:} \asprule{~light\_on :- power\_available, switch\_flipped.}
\item \textit{Fact:} 
 \asprule{~power\_available.},~
 \asprule{~switch\_flipped.}
\end{itemize}

Given these statements, the ASP solver determines the unique answer set containing \asprule{light\_on}, as all conditions in the rule body are satisfied. If we instead had \asprule{not switch\_flipped.}, the solver would exclude \asprule{light\_on} from the answer set. 

By exhaustively computing all stable models that meet the specified constraints, the ASP solver identifies all valid visualization configurations implied by our ambiguity-aware visualization tree. This systematic resolution is key to generating a complete set of valid visualizations from an ambiguous query.

\stitle{ASP Rules for Resolving Ambiguity-aware Visualization Tree.}
We formalize the visualization design space using ASP by converting each node in the ambiguity-aware visualization tree $T'$ into ASP rules. As defined in Section~\ref{sub:step1}, each node in the visualization tree is represented as a tuple $(type, operation, parameters)$, which is mapped into ASP entities, (\eg like \asprule{~entity(E, \_, \_).}) and their associated attributes (\eg like \asprule{~attribute(A, \_, \_).}.

\etitle{Rules for Explicit Nodes.} Nodes that directly specify a visualization component are encoded as entities with fully defined attributes. For example, a node indicating a specific mark selection, such as a bar chart, is encoded in ASP as:
\begin{itemize}
    \item \asprule{~entity(mark, parent\_id, mark\_id).} 
    \item \asprule{~attribute((mark, type), mark\_id, bar).}
\end{itemize}

These rules explicitly assert that the mark type is``bar''.

\etitle{Rules for Ambiguous Nodes.} Nodes that allow multiple valid interpretations are encoded using ASP choice rules.
For example, if an encoding node can correspond to either ``temp\_max'' or ``temp\_min'', we encode this ambiguity as follows:

\begin{itemize}
    \item \asprule{~1 \{ attribute((encoding, field), e\_id, temp\_max); attribute((encoding, field), e\_id, temp\_min) \}.} ensures at least one option should be selected.
    \item An accompanying integrity constraint ensures that only one of the two options is selected: \asprule{~:- attribute((encoding, field), e\_id, temp\_max), attribute((encoding, field), e\_id, temp\_min).}
\end{itemize}

This formulation forces the solver to choose exactly one interpretation for each ambiguous node.

\etitle{Rules for Implicit Nodes.} Implicit nodes represent necessary components that are not explicitly specified in the query. These nodes are encoded using placeholder attributes to indicate that the value is not determined. For example, a mark node with an unspecified chart type is represented as:

\begin{itemize}
    \item \asprule{~entity(mark, parent\_id, mark\_id).} 
    \item \asprule{~attribute((mark, type), mark\_id, \_).}
\end{itemize}

This indicates the mark exists, but its type is undetermined.

To capture the complete visualization design space, we also encode comprehensive design knowledge as ASP rules~\cite{taskvisDSE,moritz2018draco,yang2023draco2}, which fall into three categories:

\etitle{Definition Rules for Visualization.} Declarative statements that establish foundational visualization elements, such as available chart types or encoding channels. For example, \asprule{~domain((mark, type),(point;~bar;~pie)).} defines that the mark type for a chart can be point, bar, or pie.

\etitle{Hard Constraints for Visualization.} 
Mandatory conditions that any valid visualization must satisfy. For example,  the constraint 
\asprule{~violation(no\_encodings) :- entity(mark,\_,M), not entity(encoding,M,\_).} ensures that every mark has at least one visual encoding channel.

\etitle{Choice Rules for Visualization.}
Rules that govern the selection among multiple options when constructing a visualization. For example \asprule{~0 \{ attribute((encoding, field), E, N): domain((field, name), N) \} 1 :- entity(encoding,\_, E).} ensures that each encoding is associated with at most one field.

\stitle{Applying ASP Solver to Reason Valid Visualization.}
By encoding the ambiguity-aware visualization tree structure and design principles as ASP rules, we create a powerful mechanism to resolve ambiguities. The ASP solver explores all possible resolutions for ambiguous nodes, ensuring that only solutions adhering to the visualization grammar constraints are accepted. This results in a diverse set of valid visualizations, with variations in chart type, encoding mappings, and data transformations, while staying true to the original ambiguous query.

\subsection{Step 3: Ambiguous Text Query Synthesis}
\label{sub:step3}

As shown in Figure~\ref{fig:pipeline} (c), this step runs in parallel with the valid visualization synthesis described in Section~\ref{sub:step2}. Building on the ambiguity-aware visualization tree $T'$, this step aims to synthesize a corresponding ambiguous natural language query $q$.

\stitle{Task Description.}
Given the input ambiguous visualization tree $T' = \{\mathbf{A} \mid \mathbf{A} = [a_1, a_2, \ldots, a_h] \}$, the corresponding natural language query $q$ is generated using the mapping function $\mathcal{M}$:

\begin{equation}
Q = \mathcal{M}(T') = [\mathcal{M}(a_1), \mathcal{M}(a_2), \dots, \mathcal{M}(a_h)]
\label{eq:goals}
\end{equation}

where the tuple of each visualization construction action $a_i$ in $T'$ is mapped to a corresponding natural language expression $\mathcal{M}(a_i)$. 

For a given $T'$, its corresponding $q$ must satisfy the following conditions to ensure correctness:

\begin{itemize}
    \item \textit{Completeness}: Ensure that all actions in the original $T'$ are covered in the generated $q$:
    \begin{equation}
    \forall a_i \in T', \exists \mathcal{M}(a_i) \in Q
    \label{eq:completeness}
    \end{equation}

    \item \textit{Type Preservation}: $q$ must preserve the ambiguity types of the original action nodes:
    \begin{equation}
    \tau(\mathcal{M}(a_i)) = \tau(a_i), \quad \forall a_i \in T'
    \label{eq:type_preservation}
    \end{equation}
    where $\tau(a_i)$ is the ambiguity type of action node $a_i$.

    \item \textit{Boundedness}: $q$ should not introduce any actions outside of $T'$:
    \begin{equation}
    \forall \text{ expression } e \in Q, \exists a_i \in T' : e = \mathcal{M}(a_i)
    \label{eq:boundedness}
    \end{equation}
\end{itemize}

\stitle{Solution Overview.}
We leverage an LLM-based {\tt Text Query Generator} to integrate the ambiguities introduced in $T'$ into a single and coherent query $q$, ensuring that the generated query faithfully reflects all the intended ambiguous components. Finally, an {\tt Text Query Verifier} is employed to validate that $q$ accurately captures the ambiguity without introducing any extraneous semantics. This two-step process—generation followed by verification—ensures that the final query remains consistent with the design decisions encoded in $T'$ while meeting the criteria of completeness, type preservation, and boundedness.

\stitle{Text Query Diversity in Generation.}
\nlv~\cite{nlv2021} defines several distinct categories of natural language utterances---question, command,  query, and other. Since ``query'' somewhat overlaps with other styles, we focus on three main types: question, command, and caption, each representing a distinct style of user input:

\begin{itemize}
    \item \textit{Question}: Typically begins with a question word (\eg ``What'', ``How much'', ``How many'', etc.).
    \item \textit{Command}: Usually an imperative sentence (\eg ``Show a bar chart of sales by region'').
    \item \textit{Caption}: Includes non-standard phrases, incomplete sentences, or informal text conveying user intent, often brief (\eg ``SUM (Sales) vs Date'' or ``budget over time'').
\end{itemize}

To ensure diversity of the generated queries, we provide specific Text Query styles and corresponding example queries as input to the language model. These examples are randomly sampled from a large corpus to ensure variability.

\stitle{Text Query Generator.}
To systematically align the structured visualization tree with diverse natural language expressions, we define explicit input-output mappings. The input to the LLM (GPT-4o-mini-turbo) delivers essential context, including data schema, sample data, action sequences, and style requirements. This aims to ensure that the output text query: maintains linguistic grounding for all actions~\eqref{eq:completeness}, preserves ambiguity types during translation~\eqref{eq:type_preservation}, and avoids introducing any extraneous semantics~\eqref{eq:boundedness}. The complete prompt format is in Table~\ref{tab:nl_generation_prompt}.


\stitle{Text Query Verifier.}
As indicated by recent studies~\cite{jiang-etal-2024-large,weng-etal-2023-large}, LLMs outputs still require verification, particularly concerning \textit{boundedness}~\eqref{eq:boundedness}.
The verification can be performed by LLMs or human evaluators. In our preliminary experiments, we found that LLM-based verification is sufficient to achieve an accuracy of 99\%. Thus, we designed the following prompt for LLM verification as shown in Figure~\ref{tab:step_reasoning_prompt}

If $L_1$ fully covers all nodes in $T$ while $L_2$ remains empty, the $q$ is considered valid and added to the dataset. Otherwise, $q$ is classified as invalid, and it would be regenerated by the \textbf{Text Query Generator}.
This approach checks for \textit{completeness}~\eqref{eq:completeness}, \textit{type preservation}~\eqref{eq:type_preservation}, and \textit{boundedness}~\eqref{eq:boundedness}. If the verification fails, the system can regenerate the query or suggest corrections.

\begin{table}[t!]
\centering
\small \sf
\caption{Chart types, visual channels, and analytic tasks with compatible data types: $C$=Categorical, $Q$=Quantitative, $T$=Temporal, $\emptyset$=N/A.}
\label{tab:task}
\begin{tabular}{@{}ccc@{}}
\toprule
\textbf{Chart Type} & \multicolumn{1}{c}{\textbf{\begin{tabular}[c]{@{}c@{}} Encoding Channel\\ x|y|color|size|theta\end{tabular}}} & \multicolumn{1}{c}{\textbf{Analytic Task}}                            \\ \midrule
Bar &     $\{C,Q,T\}|Q|C|\emptyset|\emptyset$   & Trend, Distribution  \\ \midrule
Line &     $\{C,Q,T\}|Q|C|\emptyset|\emptyset$    & Trend, Distribution                                                     \\ \midrule
Pie  &    $\emptyset|\emptyset|C|\emptyset|Q$    & Distribution                                                 \\ \midrule
Scatter &   $Q|Q|C|Q|\emptyset$  & Correlation                                                  \\ \midrule
Heatmap &    $\{C,Q,T\}|\{C,Q\}|Q|\emptyset|\emptyset$  & Correlation                                                  \\ \midrule
Boxplot   &    $\{C\}|Q|C|\emptyset|\emptyset$   & Distribution                                                 \\ \bottomrule
\end{tabular}
\end{table}

\subsection{Step 4: Ambiguity-resolved Reasoning Path}
\label{sub:step4}

Based on the previous discussion, we have reformulated the \nlvis problem from a direct mapping $q \rightarrow \textbf{v} = \{v_1, \dots v_k \}$ to a structured process $q \rightarrow T' \rightarrow T^*_\textbf{v} \rightarrow \textbf{v}$. 
To mimic human-like reasoning workflow for ambiguity resolution, we propose decomposing the ambiguity-aware visualization generation process into a sequential reasoning path with five distinct steps, as illustrated in Figure~\ref{fig:motivation_example_1}:

\begin{equation}
q \xrightarrow{\phi_1} S_1 \xrightarrow{\phi_2} S_2 \xrightarrow{\phi_3} S_3 \xrightarrow{\phi_4} S_4 \xrightarrow{\phi_5} \textbf{v}
\end{equation}

where each $\phi_i$ represents a reasoning function and each $S_i$ represents the intermediate state after applying the corresponding reasoning function.

\stitle{Step-\ding{172}: Data Selection Reasoning.} 
The first step parses the natural language query $q$ into data components from the data table:

\begin{equation}
\phi_1(q) \rightarrow S_1 = \{a^c_1, a^c_2, \ldots, a^c_m\}
\end{equation}

where each $a^c_i$ represents a data component selection action, including column selection, value selection, and filter condition specification. The outcomes of this step correspond to the SELECT and FILTER nodes in the visualization tree (see Figure~\ref{fig:ambiguity_tree}).

\stitle{Step-\ding{173}: Chart Type Reasoning.}
The second step determines appropriate visualization mark types based on the analytic task:

\begin{equation}
\phi_2(S_1, q) \rightarrow S_2 = S_1 \cup \{a^v_1, a^v_2, \ldots, a^v_n\}
\end{equation}

where each $a^v_i$ represents a visualization design action, including analytic task identification and chart type selection. 
As shown in Table~\ref{tab:task}, existing visualization design principles~\cite{taskvis, Narechania2020,haichart} can establish a mapping relationship between tasks and chart types~\cite{taskvis, Narechania2020, coinsights}. When the text query $q$ does not explicitly specify a chart type, the identified task can guide inference, though this may introduce ambiguity as multiple chart types may be suitable for a given task. In addition, certain tasks influence encoding channel selection in the next step.

\stitle{Step-\ding{174}: Channel Mapping Reasoning.}
The third step establishes the mappings between data components and encoding channels:

\begin{equation}
\phi_3(S_2) \rightarrow S_3 = S_2 \cup \{a^m_1, a^m_2, \ldots, a^m_p\}
\end{equation}

where each $a^m_i$ represents a channel mapping action, such as assigning data columns to encoding channels like X, Y, color, or size. This step ensures that data columns are mapped appropriately, aligning with visualization design principles, where some mapping relationships are shown in Table~\ref{tab:task}.

\stitle{Step-\ding{175}: Data Transformation Reasoning.}
The fourth step specifies necessary data transformations based on the channel mappings:

\begin{equation}
\phi_4(S_3) \rightarrow S_4 = S_3 \cup \{a^t_1, a^t_2, \ldots, a^t_r\}
\end{equation}

where each $a^t_i$ represents a data transformation action, including aggregation, binning, sorting, and filtering operations, these transformations prepare the data to be properly visualized according to the selected chart type and channel mappings. 

\stitle{Step-\ding{176}: Visualization Synthesis Reasoning.}
The final step is to integrate all reasoning steps to generate a set of valid visualizations:

\begin{equation}
\phi_5(S_4) \rightarrow \textbf{v} = \{v_1, v_2, \ldots, v_k\}
\end{equation}

where each $v_i$ represents a valid visualization specification.
This process can produce multiple valid visualizations that address different aspects of the ambiguity in the original query (see Figure~\ref{fig:motivation_example_1}).

The four reasoning steps (Step-\ding{172} to Step-\ding{175}) outlined in the ambiguity-resolved reasoning path are not strictly bound by a fixed sequence and can be executed in any order, provided all steps are completed before the final visualization synthesis (Step-\ding{176}). This flexibility arises because each step addresses a distinct aspect of the visualization process—data selection, chart type reasoning, channel mapping, and data transformation—and their interdependencies are managed through the shared ambiguity-aware visualization tree $T'$. The exact order may vary depending on the text query; for example, if the query lacks any clues about the chart type, chart type reasoning may occur last, after all selected data and possible channel mappings have been considered.

This structured reasoning process systematically addresses ambiguity at each step while adhering to visualization design principles. Each step builds upon prior decisions, progressively refining the visualization specifications to account for multiple valid interpretations of the original text query.

Formally, the complete reasoning path can be expressed as the composition of the step-wise reasoning functions:

\begin{equation}
\mathcal{F}(q, D) = (\phi_5 \circ \phi_4 \circ \phi_3 \circ \phi_2 \circ \phi_1)(q, D) \rightarrow \textbf{v}
\label{eq:reverse}
\end{equation}

This decomposition simplifies the ambiguity-aware \nlvis process, breaking down complex reasoning into steps that better align with LLMs' strengths in natural language understanding and generation.  
Techniques like chain-of-thought prompting or step-wise direct preference optimization (step-DPO)~\cite{StepDPO2024, LetsVS2023} can further improve LLM performance. 

Finally, as shown in Figure~\ref{fig:pipeline} (d), the LLM-based step-wise reasoning generator takes the text query \( q \), the generated unambiguous visualization \( v \), and the ambiguous visualization tree \( T' \) as input. It then performs reverse reasoning for each step~\eqref{eq:reverse}, generating text-based reasoning descriptions. 
For example, when resolving chart type ambiguity in Figure~\ref{fig:motivation_example_1}, the LLM reasons, ``Since this query requests a trend analysis over time, either bar charts or line charts would be appropriate, as both effectively represent temporal patterns in the data'' for Step-\ding{173}. The complete prompt format is in Table~\ref{tab:step_reasoning_prompt}.


\section{More Details of \dataset}
\label{apdx:dataset}

\subsection{Detailed Example in \dataset}
\begin{figure*}[t!]
	\centering
    \vspace{-1.em}
\includegraphics[width=\textwidth]{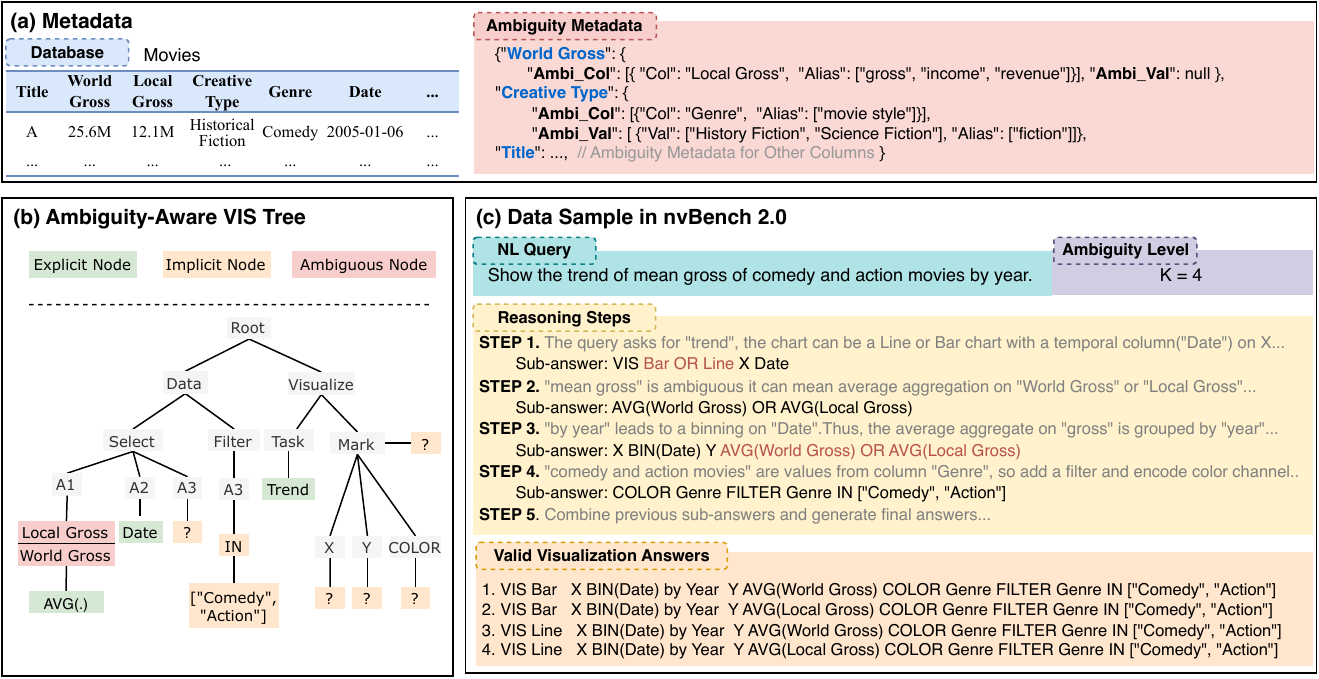}
\vspace{-2em}
	\caption{An example in \data.}
	\label{fig:data_sample}
\end{figure*}

Figure~\ref{fig:data_sample} illustrates an example sample in the \dataset, showcasing how the data is stored and the information it contains. Figure~\ref{fig:data_sample} (a) Presents the data schema for the "Movies" database, incorporating ambiguity metadata that highlights potential ambiguities, such as column aliases (e.g., ``World Gross'' and ``Local Gross'' both linked to ``gross'') and value aliases (e.g., ``History Fiction'' and ``Science Fiction'' both linked to ``fiction'').

Figure~\ref{fig:data_sample} (b) Displays the Ambiguity-Aware VIS Tree, depicting the hierarchical structure of ambiguous user intent, with explicit nodes shown in green, implicit nodes in yellow, and ambiguous nodes in red, revealing the underlying ambiguity intents for the data sample in (c).

Figure~\ref{fig:data_sample} (c) represents a data sample within \dataset, beginning with the text query "Show the trend of the mean gross of comedy and action movies by year" and an ambiguity level of $K=4$, demonstrating the number of gold answers available for this query. The sample includes reasoning steps, each with 1-2 sentences of logical reasoning and a sub-answer, culminating in the four gold answers.

\subsection{Detailed Statistics of \dataset}

\begin{figure}[t!]
    \centering
    \includegraphics[width=\linewidth]{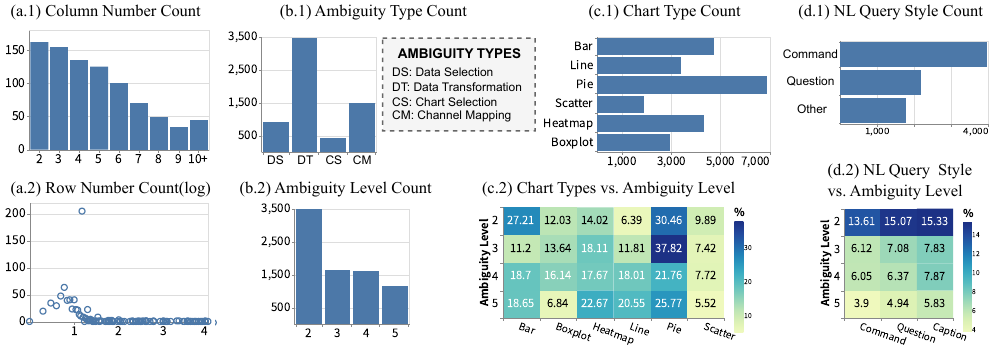}
    \vspace{-2em}
    \caption{Detailed Statistics of \data.}
    \label{fig:data_statistic_old}
\end{figure}

\stitle{Data Tables.}
Figure~\ref{fig:data_statistic_old} (a.1) shows that most tables in our dataset have 2--5 columns, with fewer than 50 tables having more than 8 columns. As Figure~\ref{fig:data_statistic_old} (a.2) illustrates (log scale), row counts range widely, from 10--1000 rows for many tables to outliers exceeding 10,000 rows. This variety ensures that \dataset tests system performance across both small and large datasets.
 
\stitle{Ambiguity Types and Levels.}
An important contribution of \dataset is the systematic introduction of controlled ambiguity levels. 
Figure~\ref{fig:data_statistic_old} (b.1) categorizes ambiguity by type: Data Transformation (DT) ambiguities are most prevalent ($\sim$ 3,500 examples), followed by Channel Mapping (CM) ambiguities ($\sim$ 1,500 examples), with Data Selection (DS) and Chart Type Selection (CT) ambiguities represented by approximately 900 and 400 examples, respectively.
As shown in Figure~\ref{fig:data_statistic_old} (b.2), the majority of samples (approximately 3,500) have an ambiguity level of 2, indicating that two valid visualizations exist. The dataset also contains a substantial number of samples with ambiguity levels of 3, 4, and 5, enabling a thorough evaluation of systems under increasingly complex ambiguous scenarios. 
Figure~\ref{fig:data_statistic_old} (c.2) and (d.2) further illustrates the relationship between ambiguity levels and two factors: chart types (c.2) and NL styles (d.2), showing comprehensive data coverage.

\stitle{Visualizations.}
Figure~\ref{fig:data_statistic_old} (c.1) shows the distribution of chart types in \dataset. Pie charts are the most common, with around 6,000 examples, followed by bar charts ($\sim$4,000) and heatmaps ($\sim$3,500). Additionally, line charts ($\sim$2,800), boxplots ($\sim$2,000), and scatter plots ($\sim$1,500) are also well-represented, ensuring that the benchmark covers all major visualization types. This distribution reflects common visualization practices, where pie and bar charts are widely used for categorical comparisons, while the other types serve specialized analytical needs.

\stitle{Text Queries.}
Figure~\ref{fig:data_statistic_old} (d.1) presents the natural language query distribution. Command-based queries (\eg ``Show me the sales by region'') are most frequent ($\sim$4,000). Question-based queries  (\eg ``What are the sales trends?'') and caption-like statements (\eg ``SUM (Sales) vs Date'') appear in about 2,000 and 1,800 instances, respectively. 
Table~\ref{tab:nl_style} provides a detailed breakdown of NL styles across different chart types, along with word count statistics. 
Commands, questions, and captions are distributed across various chart types, with pie charts receiving the highest number of queries (6,872). 
The average word count remains consistent ($\sim$14 words), with captions exhibiting the longest maximum length (65 words). This distribution highlights the dataset’s diversity in both linguistic structure and visualization needs, ensuring that \dataset can effectively evaluate systems' capabilities to handle diverse user interactions. 

\begin{table}[t!]
\centering
\caption{Distribution of natural language styles across chart types and word count statistics}
\small
\setlength{\tabcolsep}{0.9pt} 
\renewcommand\arraystretch{0.9}
\begin{tabular}{cccccccccccc}
\toprule
\multirow{2}{*}[-1pt]{\makebox[1cm][c]{\textbf{NL Style}}} &
  \multicolumn{6}{c}{\textbf{Count by Chart Type}} &
  \multirow{2}{*}{\textbf{Total}} &
  \multicolumn{3}{c}{\textbf{Word Count}} \\ 
\cmidrule(lr){2-7} \cmidrule(lr){9-11}
         & Bar  & Line & Pie  & Scatter & Boxplot & Heatmap &      & Avg.  & Max & Min \\ 
\midrule
Command  & 1368 & 922  & 1922 & 608     & 1319    & 894     & 2338 & 14.20  & 60 & 6 \\
Question & 1570 & 1084 & 2299 & 679     & 1403    & 966     & 2636 & 14.04 & 39 & 5 \\
Caption  & 1779 & 1363 & 2651 & 581     & 1589    & 1079    & 2904 & 14.00 & 65 & 5 \\
\midrule
Total    & 4717 & 3369 & 6872 & 1868    & 4311    & 2939    & 7878 & 14.07 & 65  & 5 \\
\bottomrule
\end{tabular}
\label{tab:nl_style}
\end{table}

\section{More Details of Experimental setups}
\label{apdx:exp}

\stitle{Methods.}
We evaluate the performance on ambiguous \nlvis tasks using both prompting-based and fine-tuning-based methods with our \data. The primary goal is to assess the model's ability to generate diverse and semantically accurate visualizations in response to ambiguous \nlq queries.

\etitle{Prompting-based Methods.} 
We evaluate two prompting strategies with GPT-4o-mini, GPT-4o and Qwen2.5-7B-Instruct model:

\begin{itemize}
    \item \textbf{Direct Prompting:} See Table~\ref{tab:basic_experiments_prompt} for complete prompt structure. The model receives structured \textit{Data Information} and an \textit{Text Query} as input, generating 1-5 distinct charts to cover possible interpretations of ambiguous queries. 
    
    \item \textbf{Step Prompting:} See Table~\ref{tab:step_exp_prompt} for complete prompt structure. Models are guided to ``\texttt{think step-by-step}'', explicitly articulating their reasoning process before generating visualizations. Models using this approach are denoted with a ``-Step'' suffix.
\end{itemize}

\etitle{Supervised Fine-tuning Method.}

\begin{itemize}
    \item \textbf{Qwen2.5-7B-SFT}: We performed supervised fine-tuning on the Qwen2.5-7B-Instruct model using the training set, enabling direct generation of multiple Vega-Lite definitions without step-wise reasoning. Training involved three epochs with a global batch size of 16, a learning rate of 2e-5, the AdamW optimizer, and a cosine learning rate scheduler with a 0.1 warmup ratio.
\end{itemize}

\etitle{Preference Learning Method.}

\begin{itemize}
    \item \textbf{\model}: We designed \model to handle the ambiguity in \nlvis through step-wise reasoning as detailed in Section~\ref{sec:model}. After the initial supervised fine-tuning of Qwen-2.5-7B-Instruct, we constructed a preference dataset from the \data development set for Step-DPO training. This process used one epoch with a global batch size of 4, a linearly decaying learning rate from 2e-6, and the AdamW optimizer.
\end{itemize}

\stitle{Evaluation Metrics.} Detailed explanation for evaluation metrics:

\begin{itemize}
    
    \item \textbf{Precision@K (P@K)}: Assesses recommendation accuracy by calculating the proportion of valid visualizations among the top-K outputs. Higher P@K indicates more trustworthy recommendations, with fewer incorrect visualizations shown to users.
    
    \item \textbf{Recall@K (R@K)}: Quantifies how completely the model covers the golden visualization space by measuring the proportion of valid visualizations successfully identified. This captures the model's ability to represent multiple valid interpretations for ambiguous queries.
    
    \item \textbf{F1@K}: Provides a balanced measure that combines precision and recall through their harmonic mean. This comprehensive metric rewards systems that achieve both high coverage of the golden answer space and high accuracy in their recommendations.
\end{itemize}

For all experiments, we report these metrics at $K$ $\in$ \{1, 3, 5\} to evaluate performance across different recommendation set sizes.






\section{More Details of Error Analysis}
\label{apdx:error}

\begin{figure}[p]  
  \centering  
  \includegraphics[width=\columnwidth]{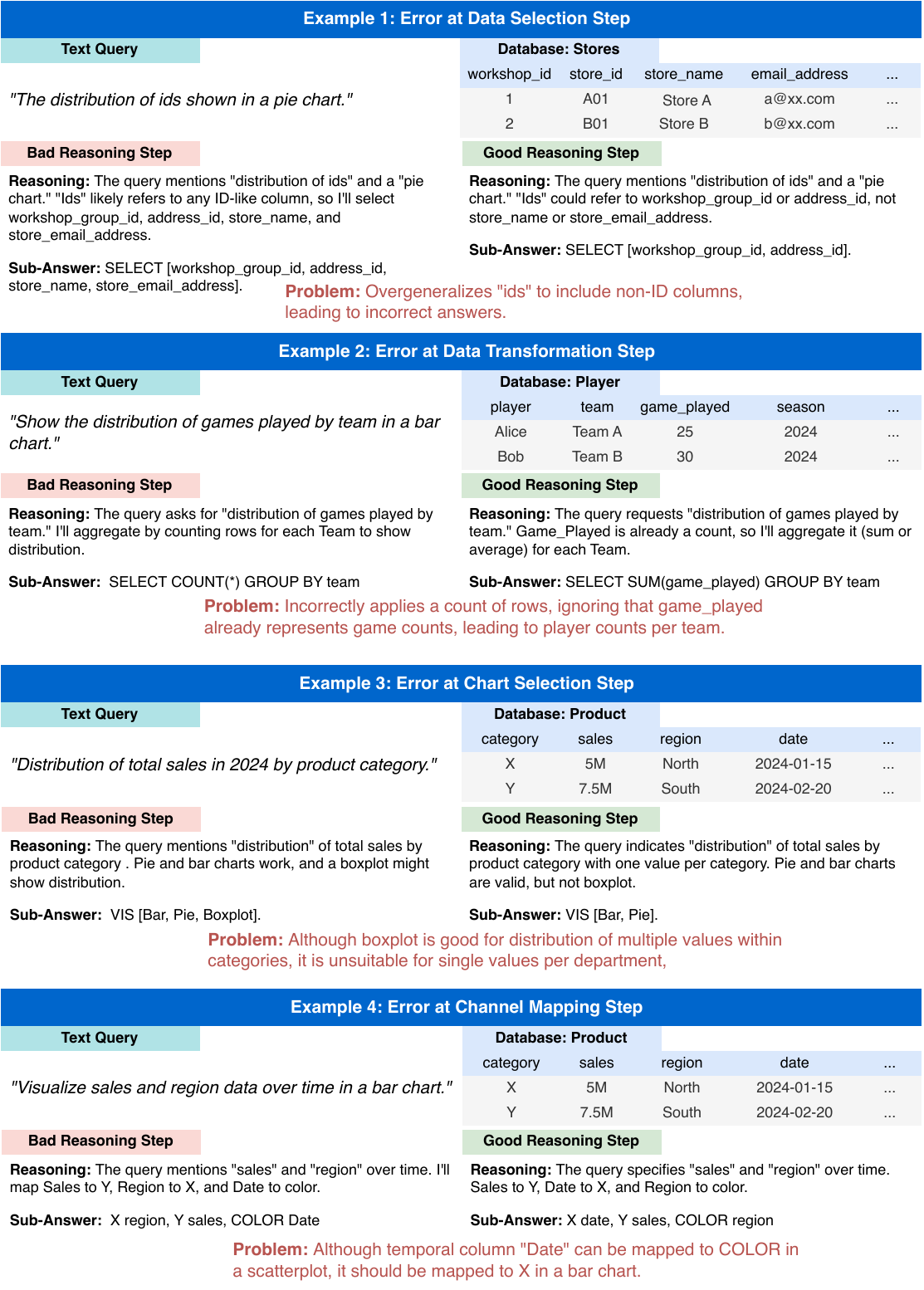}  
  \caption{Examples of Stepwise Reasoning Errors in the nvBench 2.0 dataset, highlighting common pitfalls in the Text-to-Visualization process.}  
  \label{fig:step-reasoning-error}  
   \vspace{-1em}
\end{figure}  

\subsection{Qualitative Error Analysis in Reasoning Paths.}

Figure~\ref{fig:step-reasoning-error} outlines frequent errors in the stepwise reasoning process for ambiguous Text2VIS tasks across four key steps:

\begin{itemize}
    \item At the Data Selection step, errors frequently arise when models overgeneralize column references, selecting irrelevant or overly broad sets of data that do not align with the query's intent.
    \item During the Data Transformation step, mistakes often occur due to misinterpretation of aggregation requirements, such as applying incorrect operations like counting rows instead of summing specific values.
    \item At the Chart Selection step, errors commonly stem from choosing inappropriate chart types that fail to match the analytical task or data characteristics, leading to ineffective visualizations.
    \item In the Channel Mapping step, issues frequently emerge from incorrect assignments of data to visualization channels, such as mapping a temporal field to a color channel instead of an axis, resulting in misleading representations.
\end{itemize}

The stepwise reasoning step information provided in these examples is valuable for further ambiguous \nlvis system development, as it offers insights into common reasoning pitfalls, enabling the design of more robust models that can learn from and correct these errors to improve accuracy and interpretability in handling ambiguous queries.



\section{Limitations}
\label{apdx:limit}

While \dataset introduces significant advancements in ambiguity-aware Text2VIS benchmarking, several limitations remain that present opportunities for future research:

\textbf{Limited Coverage of Visualization-Adjacent Tasks:} Although our benchmark focuses on Text2VIS ambiguity resolution, it does not extend to related domains such as Text2SQL. The step-wise reasoning approach could potentially be adapted to handle SQL generation with ambiguous queries, particularly since visualization and database queries share similar data operations. Future work could explore the integration of both Text2VIS and Text2SQL ambiguity resolution within a unified framework.

\textbf{Restricted Chart Types and Components:} Though nvBench 2.0 includes six chart types (bar, line, pie, scatter, heatmap, and boxplot), it does not cover the full spectrum of visualization techniques. Advanced chart types like treemaps, network diagrams, geographic maps, and multi-view coordinated visualizations are not included. Additionally, the benchmark lacks support for more sophisticated chart components such as error bars, trend lines, annotations, interactive elements, and customizable legends that are often crucial for comprehensive data storytelling.

\textbf{Limited Integration with Statistical Analysis:} The current benchmark treats visualization as the primary goal rather than integrating it with deeper statistical analysis intents. Users often request visualizations to support specific analytical objectives (hypothesis testing, correlation analysis, outlier detection, clustering) that require a tighter coupling between visualization and statistical computation. Future work could expand the benchmark to include cases where visualization serves as a component within broader analytical workflows.

\textbf{Absence of Conversational Context:} nvBench 2.0 evaluates standalone queries without considering the conversational context in which they might appear. In real-world scenarios, visualization requests often occur within multi-turn dialogues where context from previous exchanges influences interpretation. The benchmark does not account for these contextual dependencies, limiting its ability to evaluate systems in realistic interactive settings where ambiguity resolution might span multiple conversational turns.
\section{Ethic Statement}
\label{apdx:ethic}

\begin{table}[t]
\centering
\caption{Licenses List for Assets Used}
\label{tab:licenses}
\begin{tabular}{lll}
\toprule
\textbf{Asset} & \textbf{Usage} & \textbf{License} \\
\midrule
GPT-4o~\cite{openai2024gpt4technicalreport} & Baselines and query verification & \href{https://openai.com/policies/terms-of-use/}{Custom License} \\
GPT-4o-mini~ &  Baselines, metadata generation, query synthesis & \href{https://openai.com/policies/terms-of-use/}{Custom License} \\
Qwen2.5-7B~\cite{bai2023qwentechnicalreport} & Baselines and model fine-tuning & Apache-2.0 \\
nvBench Dataset~\cite{nvBench2021} & Source for data tables and seed visualizations & MIT \\
BIRD Dataset~\cite{BIRD2023} & Source for additional data tables & CC BY-SA 4.0 \\
ConceptNet~\cite{conceptNet} & Semantic alias discovery in ambiguity metadata & CC BY-SA 4.0 \\
ASP Solver (Clingo)~\cite{gebser2019multi} & Visualization query resolution & MIT \\
\bottomrule
\end{tabular}
\end{table}

This paper introduces \dataset, a novel benchmark for ambiguous Text-to-Visualization tasks, and evaluates the capabilities of LLMs in resolving visualization ambiguity. Our work is not intended to provoke anxiety, but rather to gain a better understanding of how LLMs can be leveraged to interpret ambiguous natural language queries for data visualization. This study aims to foster discussions on how visualization systems can better accommodate the inherent ambiguity in human requests, and how humans can more effectively utilize LLM-powered visualization tools.

In developing \dataset, we have ensured that the dataset does not contain sensitive or personally identifiable information. The data tables used in our benchmark are derived from public datasets (nvBench~\cite{nvBench_dataset} and BIRD~\cite{BIRD2023}) and have been carefully reviewed to exclude potentially harmful or private content.

In our experimental evaluations involving human reviewers for query verification, participants received appropriate compensation and ensured adequate rest periods between review sessions. We rigorously protect participants' personal information, ensuring that their information remains confidential and is not disclosed in this document or the GitHub repository.

Our source code and data are under GPL-3 license, and we follow the licenses of assets used in this paper, as listed in Table~\ref{tab:licenses}.


\begin{table}[t!] 
\caption{Prompt Structure for Ambiguous Text Query Synthesis} 
\label{tab:nl_generation_prompt} 
\centering
\fontsize{9pt}{11pt}\selectfont 

\begin{tabularx}{\textwidth}{>{\RaggedRight\arraybackslash}X} 
\noalign{\hrule height 1pt} 
\multicolumn{1}{c}{\textbf{Prompt Structure for Ambiguous Text Query Synthesis}} \\ 
\noalign{\hrule height 1pt} 

\textbf{\#\#\# Task Description:}\\
You are an intelligent assistant. You will create three text queries (command, question, and caption) based on a given data schema and action list. Each query must incorporate all information from the action list without introducing extra elements. \\ 
\hline

\textbf{\#\#\# Process (4 steps):}\\
\textbf{\# Step 1: Interpret Visualization Type.} \eg ... \\
\textbf{\# Step 2: Rephrase Data Columns.} \eg ... \\
\textbf{\# Step 3: Rephrase Data Transformations.} \eg ... \\
\textbf{\# Step 4: Rephrase Filter Conditions.} \eg ... \\
\hline

\textbf{\#\#\# Final Answer Construction:}\\
Combine the rephrased elements from steps 1-4 to create three text queries:\\
1. Command-style: Direct instruction (\eg "Plot the total sales by month for products priced over \$50")\\
2. Question-style: Inquiry format (\eg "What was the average rating for movies released during 2020?")\\
3. Caption-style: Declarative format (\eg "Distribution of customer count by region in a pie chart.") \\
\hline

\textbf{\#\#\# Input}\\
Database: \{basename\}\\
Data Columns: \{data\_schema\}\\
Data Value Examples:\{data\_value\_example\}\\
Ambiguous Column Pairs:\{ambiguous\_pairs\}\\
Action List:\{action\_list\} \\
\hline

\noalign{\hrule height 1pt} 
\end{tabularx} 

\end{table}


\begin{table}[t!]
\caption{Prompt Structure for Text Query Verification}
\label{tab:nl_verification_prompt}
\centering
\fontsize{9pt}{11pt}\selectfont
\begin{tabularx}{\textwidth}{>{\RaggedRight\arraybackslash}X}
\noalign{\hrule height 1pt}
\multicolumn{1}{c}{\textbf{Prompt Structure for Text Query Verification}} \\
\noalign{\hrule height 1pt}

\textbf{\#\#\# Task Description:}\\
You are a smart assistant. Your job is to check if a text query for a Text-to-Visualization (Text2VIS) task is valid. The query must match all parts of the provided visualization tree (T) and follow three rules: \\
1. \textbf{Completeness}: The query must include all actions and elements from the visualization tree, such as data selections, transformations, chart types, and channel mappings. \\
2. \textbf{Type Preservation}: The query must keep the same ambiguity types (e.g., Data Transformation, Channel Mapping, Data Selection, Chart Type Selection) as defined in the visualization tree. \\
3. \textbf{Boundedness}: The query must not add extra information or elements not in the visualization tree. \\
You will get the data schema, action list, visualization tree (T), and the text query (q). Verify if the query meets all three rules. If the query is invalid, explain what is missing or extra to help fix it. \\
\hline

\textbf{\#\#\# Input:}\\
\# \textbf{Database}: [baseline] \\
\# \textbf{Data Schema}: \{data\_schema\}\\
\# \textbf{Data Value Examples}: \{data\_value\_example\}  \\
\# \textbf{Visualization Tree (T)}: \{vis\_tree\}  \\
\# \textbf{Synthesized Text Query (q)}: [text\_query] \\
\hline

\textbf{\#\#\# Output:}\\
\textbf{\# Validity}: [Valid/Invalid] \\
\textbf{\# Completeness}: [Met/Not Met] - Confirm if the query includes all actions and elements from the visualization tree (T). \\
\textbf{\# Type Preservation}: [Met/Not Met] - Confirm if the query preserves the ambiguity types as defined in the visualization tree. \\
\textbf{\# Boundedness}: [Met/Not Met] - Confirm if the query avoids adding extra information not in the visualization tree. \\
\textbf{\# Feedback (if Invalid)}: Explain what makes the query invalid, including missing elements or extra information, to guide fixing or regenerating the query. \\
\hline

\noalign{\hrule height 1pt}
\end{tabularx}
\end{table}



\begin{table}[t!] 
\caption{Prompt Structure for Step-wise Reasoning Synthesis} 
\label{tab:step_reasoning_prompt} 
\centering
\fontsize{9pt}{11pt}\selectfont 

\begin{tabularx}{\textwidth}{>{\RaggedRight\arraybackslash}X} 
\noalign{\hrule height 1pt} 
\multicolumn{1}{c}{\textbf{Prompt Structure for Step-wise Reasoning Synthesis}} \\ 
\noalign{\hrule height 1pt} 

\textbf{\#\#\# Task Description:}\\
You are a good data visualization expert. Given an ambiguous/incomplete Text Query with Data Schema, and the step-by-step answer recommending visualization charts corresponding to the ambiguous/incomplete Text Query. Then, you need to fill in the reasoning process for each step. \\ 
\hline

\textbf{\#\#\# Instructions:}\\
\textbf{\# Step 1: Data Selection.} 
Select data columns and data filters mentioned in the Text Query. If the Text Query is ambiguous and can be mapped to multiple columns, use a list to indicate all columns.\\
\textbf{\# Step 2: Data Transformation.} 
Select data transformation (operation : parameter) = (aggregate : [sum, mean, count]; bin : base; sort:[ascending, descending]) mentioned in the Text Query.\\
\textbf{\# Step 3: Chart type Selection.}
Select all valid chart types for visualization based on Text Query and data selected.
If chart type indicated in the Text Query, select from chart mark=(bar, line, arc, point, rect, boxplot).
Else if no chart type mentioned, but specific analysis task mentioned in the Text Query, inference chart type by (task:chart)=(trend:[bar,line]; distribution:[bar,arc,line,boxplot]...
Also consider if the chart type can visualize selected data.\\
\textbf{\# Step 4: Selected Column-Channel Mapping.}
Map selected data columns to encoding channels = (x, y, color, size).
You should consider basic channel mapping feasibility.
Answer with all valid chart-channel-column mapping solutions.\\
\textbf{\# Step 5: Visualization Synthesis.}
Based on previous steps, synthesis the final visualizations. \\
\hline

\textbf{\#\#\# Input:}\\
Database: \{basename\}\\
Data Columns: \{data\_schema\}\\
Data Value Examples:\{data\_value\_example\} \\
Text Query: \{text\_query\}\\
\hline

\textbf{\#\#\# Output}\\
\# Step 1: <reasoning>...</reasoning> <answer> ... <answer>\\
\# (Step 2-4) \\
\# Step 5: <answer> [ VIS 1, VIS 2, ..., VIS k ] <answer>\\

\noalign{\hrule height 1pt}
\end{tabularx} 

\end{table}

\begin{table}[t!]
\caption{Prompt Structure for Basic Experiments}
\label{tab:basic_experiments_prompt}
\centering
\fontsize{9pt}{11pt}\selectfont
\begin{tabularx}{\textwidth}{>{\RaggedRight\arraybackslash}X}
\noalign{\hrule height 1pt}
\multicolumn{1}{c}{\textbf{Prompt Structure for Basic Experiments}} \\
\noalign{\hrule height 1pt}
\textbf{\#\#\# Task Description:} \\
You are a good data visualization expert. Given an ambiguous/incomplete Natural Language Query and a Data Table, please recommend 1 to 5 different charts corresponding for the ambiguous/incomplete NL Query. Please strictly follow the output format.\\
\midrule
\textbf{\#\#\# Example:} \\
\textbf{\# Input:} \\
Database: \{basename\} \\
Data Columns: \{data\_schema\} \\
Data Value Examples: \{data\_value\_example\} \\
Query: \{text\_query\} \\
\textbf{\# Output:} \\
<answer> [ VIS 1, VIS 2, ..., VIS k ] </answer> \\
\bottomrule
\end{tabularx}
\end{table}

\begin{table}[t!]
\caption{Prompt Structure for Stepwise Reasoning Experiments}
\label{tab:step_exp_prompt}
\centering
\fontsize{9pt}{11pt}\selectfont
\begin{tabularx}{\textwidth}{>{\RaggedRight\arraybackslash}X}
\noalign{\hrule height 1pt}
\multicolumn{1}{c}{\textbf{Prompt Structure for Stepwise Reasoning Experiments}} \\
\noalign{\hrule height 1pt}
\textbf{\#\#\# Task Description:} \\
You are a good data visualization expert. Given an ambiguous/incomplete Natural Language Query and a Data Table, please recommend 1 to 5 different charts corresponding for the ambiguous/incomplete NL Query. \\
Please think step by step and strictly follow the output format.\\
\midrule
\textbf{\#\#\# Example:} \\
\textbf{\# Input:} \\
Database: \{basename\} \\
Data Columns: \{data\_schema\} \\
Data Value Examples: \{data\_value\_example\} \\
Query: \{text\_query\} \\
\textbf{\#\#\# Output:} \\
\# Step 1: <reasoning>...</reasoning> <answer> ... </answer>\\
\# (Step 2-4) \\
\# Step 5: <answer> [ VIS 1, VIS 2, ..., VIS k ] </answer>\\
\bottomrule
\end{tabularx}
\end{table}

\end{document}